\documentclass[letterpaper, 10 pt, conference]{ieeeconf}  

\IEEEoverridecommandlockouts                              

\usepackage{graphics} 
\usepackage{float} 
\usepackage{epsfig} 
\usepackage{amsmath} 
\usepackage{amssymb}  

\usepackage{algorithm}
\usepackage[noend]{algorithmic}

\usepackage[dvipsnames]{xcolor} 

\usepackage[dvips,frame,rotate,import]{xy}                   

\usepackage{graphicx}

\usepackage{url}

\newcommand{\algorithmTextSize}{}

\newcommand{\onlyInLongVersion}[1]{#1}       
\newcommand{\onlyInShortVersion}[1]{}        
\newcommand{\extraDiscussion}[1]{#1}

\newcommand{\algRaiseSize}{-.70cm}
\newcommand{\algRaiseSizeB}{.23cm}
\newcommand{\algRaiseSizeC}{-.65cm}

\title{\LARGE \bf
Multi-Agent Coverage in Urban Environments
}

\author{Shivang Patel$^1$, Senthil Hariharan$^1$, Pranav Dhulipala$^1$\\ Ming C Lin$^{1,2}$, Dinesh Manocha$^{1,2,3}$, Huan Xu$^{1,4,5}$ and Michael Otte$^{1,2,4}$
\thanks{Authors are with: 
$^{1}$Maryland Robotics Center,
$^{2}$The Department of Computer Science,
$^{3}$The Department of Electrical and Computer Engineering,
$^{4}$The Department of Aerospace Engineering, and
$^{5}$The Instiitute for Systems Research, University of Maryland, College Park.
{\tt\small spatel43@umd.edu,} {\tt\small\{lin, dm\}@cs.edu,} {\tt\small \{mumu, otte\}@umd.edu}}
\thanks{This work was supported by DARPA cooperative agreement HR00111820028 as part of DARPA OFFSET, ARO grant W911NF-19-1- 0069}
}

\begin{document}

\newcommand{\reals}{\mathbb{R}}
\newcommand{\minkowskiSum}{\oplus}
\newcommand{\concat}{\circ}
\newcommand{\argmin}{\operatornamewithlimits{arg\ min}}
\newcommand{\argmax}{\operatornamewithlimits{arg\ max}}


\newcommand{\numRobots}{n}
\newcommand{\robotTeam}{R}
\newcommand{\robotIndex}{i}
\newcommand{\otherRobotIndex}{j}
\newcommand{\robot}[1]{r_{#1}}


\newcommand{\timeVariable}{t}
\newcommand{\timeVariableMax}{t_{\mathrm{max}}}

\newcommand{\spacespacetd}[1]{\chi_{#1}^{3}}
\newcommand{\spacespace}[1]{\chi_{#1}}

\newcommand{\spacePoint}{x}
\newcommand{\teamSpacePoint}{\mathbf{x}}

\newcommand{\pdfFunction}{f_{\chi}}
\newcommand{\pdfFunctionFree}{f_{\chi_{free}}}
\newcommand{\pdfFunctionVacant}{f_{\chi_{Vacant}}}

\newcommand{\motionPath}[1]{\rho_{#1}}
\newcommand{\motionPathObserving}[1]{\tilde{\rho}_{#1}}

\newcommand{\pathset}{\psi}
\newcommand{\pathsetObserving}{\tilde{\psi}}

\newcommand{\spacecone}[1]{K({#1}, d_{s_{max}})}
\newcommand{\disc}{B}

\newcommand{\altitude}{h}
\newcommand{\optimalAltitude}{\tilde{h}}

\newcommand{\totalVisitDuration}{g_{\pathsetObserving,\timeVariableMax}}

\newcommand{\indicatorFunction}{\mathbf{1}_{\mathrm{see}}}

\newcommand{\multiPathObservingSpace}{\tilde{\Psi}}
\newcommand{\multiPathSpace}{\Psi}

\newcommand{\pathLength}{\ell}

\newcommand{\voronoiCell}{C}

\newcommand{\traj}[1]{r_{#1}}
\newcommand{\probdist}{M(X)}
\newcommand{\metric}{\phi(t)}
\newcommand{\fourcoeffc}{c_{k}(t)}
\newcommand{\fourcoeffu}{\mu_{k}}
\newcommand{\fourbase}{\f_k}

\newcommand{\subspace}[1]{\spacespace{#1}}
\newcommand{\noteMO}[2]{{\color{VioletRed}#1}{\marginpar{\footnotesize \color{BlueGreen}#2}}}


\newcommand{\boustrophedon}[1]{ \mathtt{Boustrophedon}(#1)}
\newcommand{\singleAgentErgodic}[1]{ \mathtt{singleErgodic}(#1)}
\newcommand{\singleAgentErgodicObstacles}[1]{ \mathtt{singleErgAvoidObs}(#1)}
\newcommand{\flyOverBuildings}[1]{\mathtt{flyOverBldgs}(#1)}
\newcommand{\rotateCycle}[1]{\mathtt{rotateCycle}(#1)}

\newcommand{\initializeErgodicParams}{\mathtt{initErgParams}}
\newcommand{\calculateNextStep}{\mathtt{nextStep}}
\newcommand{\updateCurrentDist}{\mathtt{updateCurrentDists}}
\newcommand{\calculateVectorField}{\mathtt{calcVectorField}}
\newcommand{\projectionOnto}{\mathtt{projectOnto}}

\newcommand{\voronoiPartition}{\mathtt{voronoiPart}}
\newcommand{\gridPartition}{\mathtt{gridPartition}}
\newcommand{\centroid}{\mathtt{centroid}}


\maketitle
\thispagestyle{empty}
\pagestyle{empty}

\begin{abstract}

We study multi-agent coverage algorithms for autonomous monitoring and patrol in urban environments. 
%
%
We consider scenarios in which a team of flying agents uses downward facing cameras (or similar sensors) to observe the environment outside of buildings at street-level. Buildings are considered obstacles that impede movement, and cameras are assumed to be ineffective above a maximum altitude. We study multi-agent urban coverage problems related to this scenario, including: (1) static multi-agent urban coverage, in which agents are expected to observe the environment from static locations, and (2) dynamic multi-agent urban coverage where agents move continuously through the environment. We experimentally evaluate six different multi-agent coverage methods, including: three types of ergodic coverage (that avoid buildings in different ways), lawn-mower sweep, voronoi region based control, and a naive grid method. We evaluate all algorithms with respect to four performance metrics (percent coverage, revist count, revist time, and the integral of area viewed over time), across four types of urban environments $[\text{low density, high density}] \times [\text{short buildings, tall buildings}]$, and for team sizes ranging from 2 to 25 agents. 
We believe this is the first extensive comparison of these methods in an urban setting.
Our results highlight how the relative performance of static and dynamic methods changes based on the ratio of team size to search area, as well the relative effects that different characteristics of urban environments (tall, short, dense, sparse, mixed) have on each algorithm.


\end{abstract}

\section{Introduction}

    \newcommand{\widthB}{2.7cm}
    \newcommand{\widthBinside}{2.8cm}
    \newcommand{\heightSpacerB}{.2cm}
    \newcommand{\widthSpacerB}{-.2cm}
    
\begin{figure}[t!]

   \centering

    Multi-Agent Coverage Algorithms We Experimentally Evaluate In Urban Environments
    
    \vspace{\heightSpacerB}

    \begin{minipage}{\widthB}
      \begin{xy}
        \xyimport(100,100){\includegraphics[width=\widthBinside]{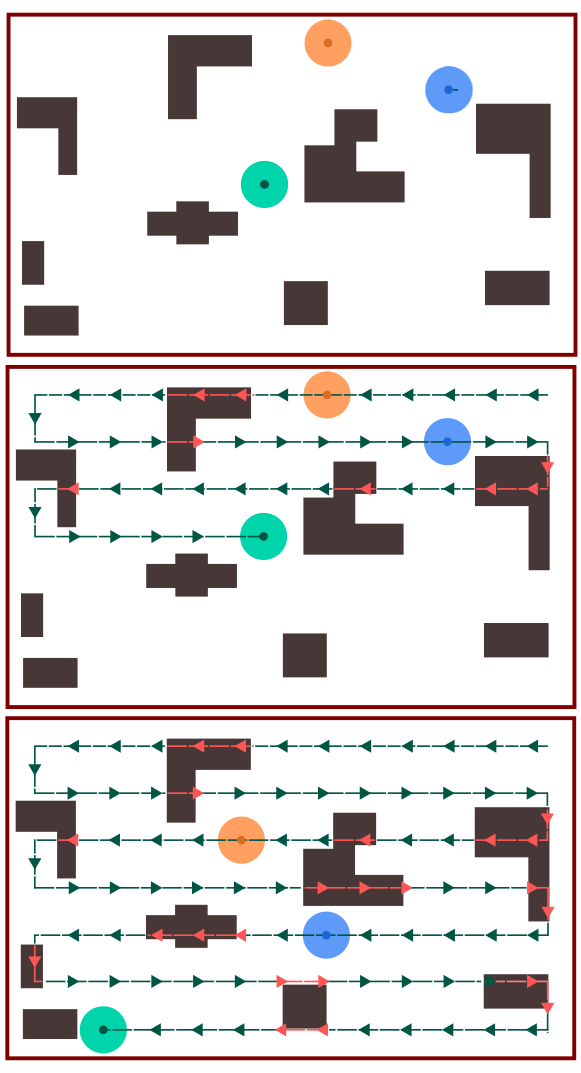}}
        ,(50,102)*{\text{\footnotesize \bf Lawnmower Sweep}}
      \end{xy}
    \end{minipage}
    \hspace{\widthSpacerB}
    \begin{minipage}{\widthB}
      \begin{xy}
        \xyimport(100,100){\includegraphics[width=\widthBinside]{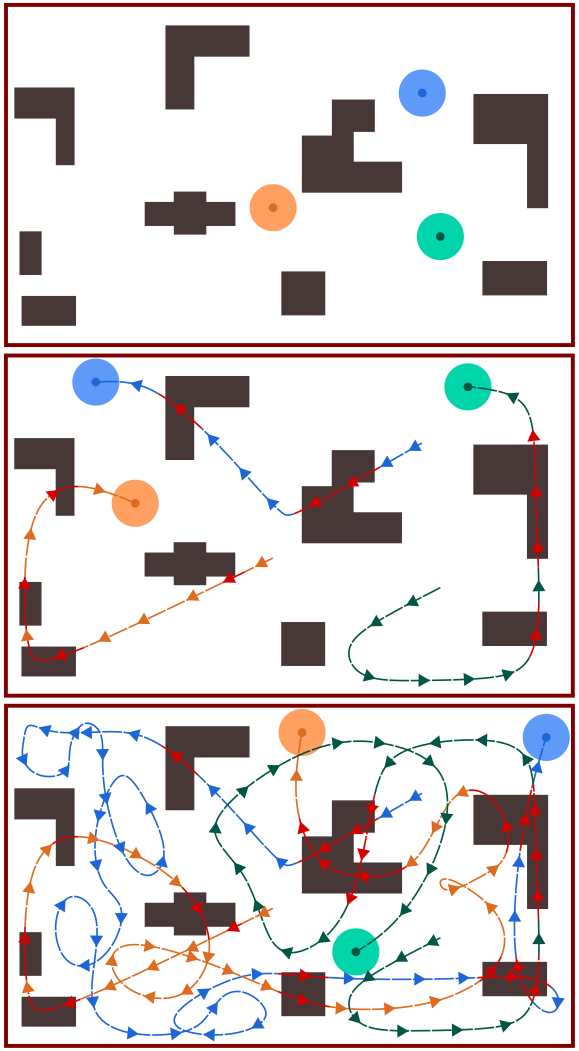}}
        ,(50,103)*{\text{\footnotesize \bf Ergodic Sweep}}
      \end{xy}
    \end{minipage}
    \hspace{\widthSpacerB}
    \begin{minipage}{\widthB}
      \begin{xy}
        \xyimport(100,100){\includegraphics[width=\widthBinside]{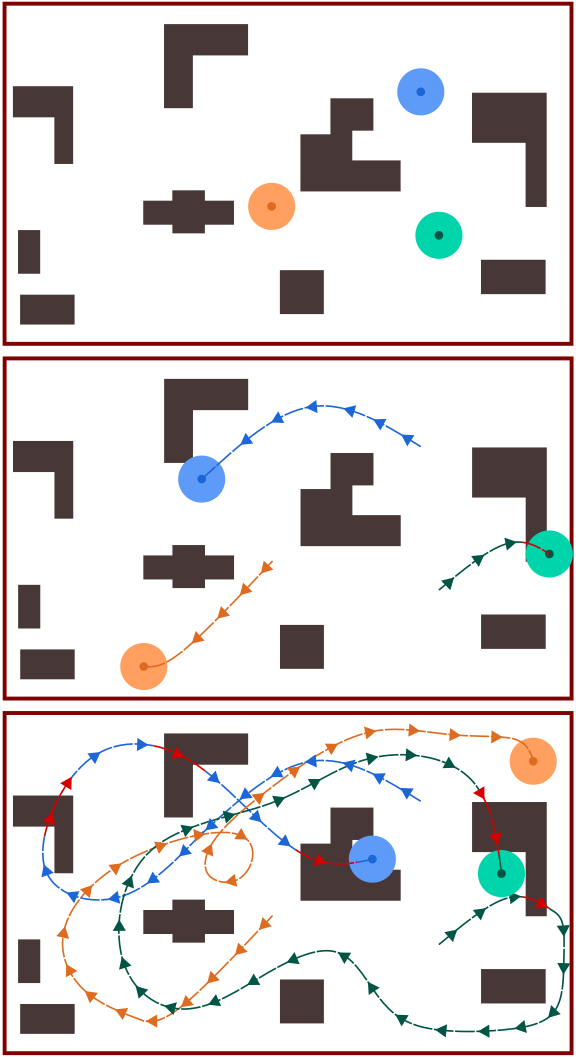}}
        ,(50,103)*{\text{\footnotesize \bf Non-Uniform Ergodic}}
      \end{xy}
    \end{minipage}

    \vspace{\heightSpacerB}

    \begin{minipage}{\widthB}
      \begin{xy}
        \xyimport(100,100){\includegraphics[width=\widthBinside]{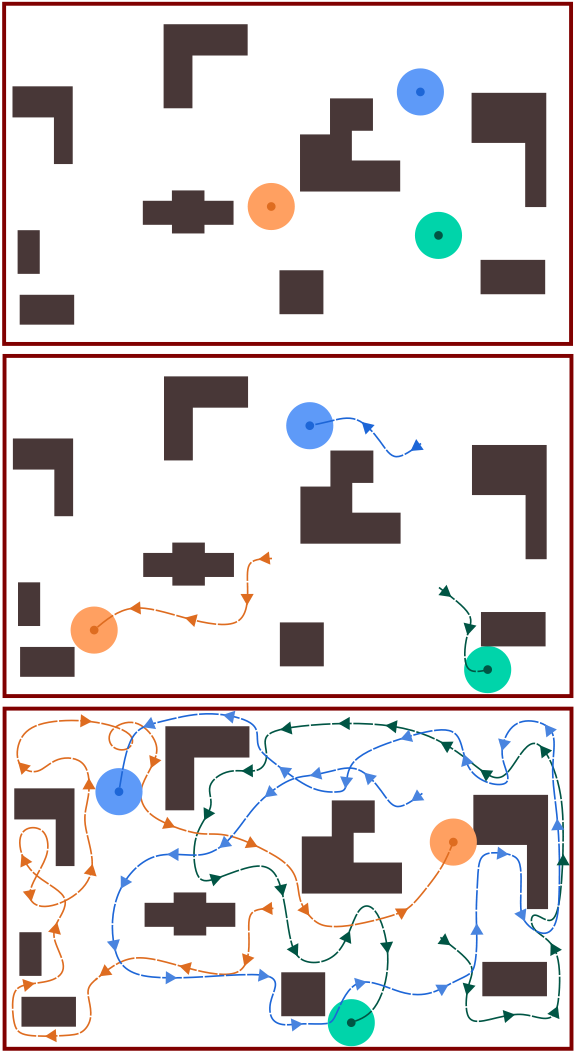}}
        ,(50,103)*{\text{\footnotesize \bf Ergodic Obstacle Avoid}}
      \end{xy}
    \end{minipage}
    \hspace{\widthSpacerB}
    \begin{minipage}{\widthB}
      \begin{xy}
        \xyimport(100,100){\includegraphics[width=\widthBinside]{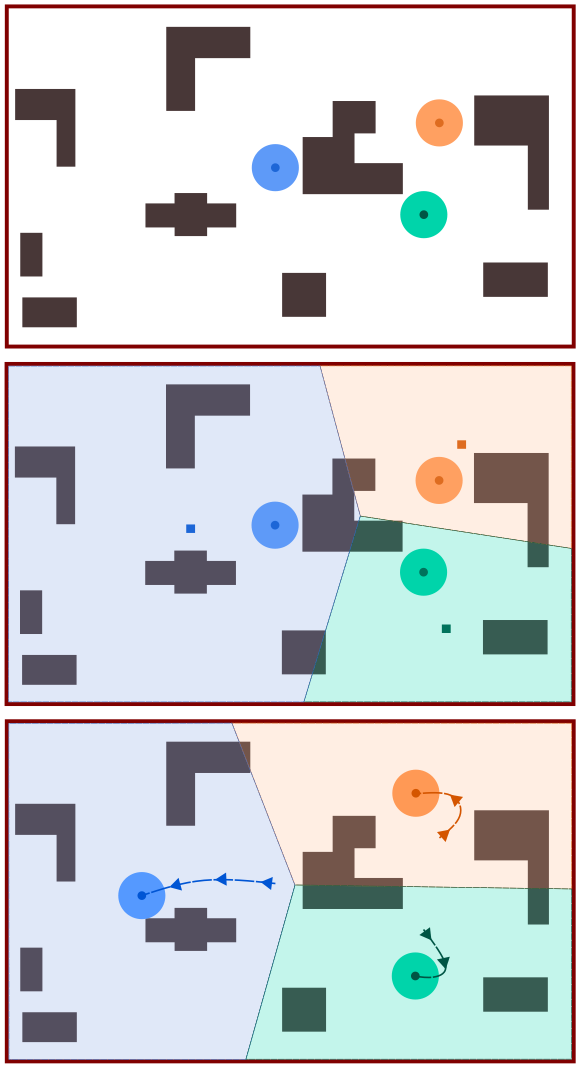}}
        ,(50,103)*{\text{\footnotesize \bf Voronoi}}
      \end{xy}
    \end{minipage}
    \hspace{\widthSpacerB}
    \begin{minipage}{\widthB}
      \begin{xy}
        \xyimport(100,100){\includegraphics[width=\widthBinside]{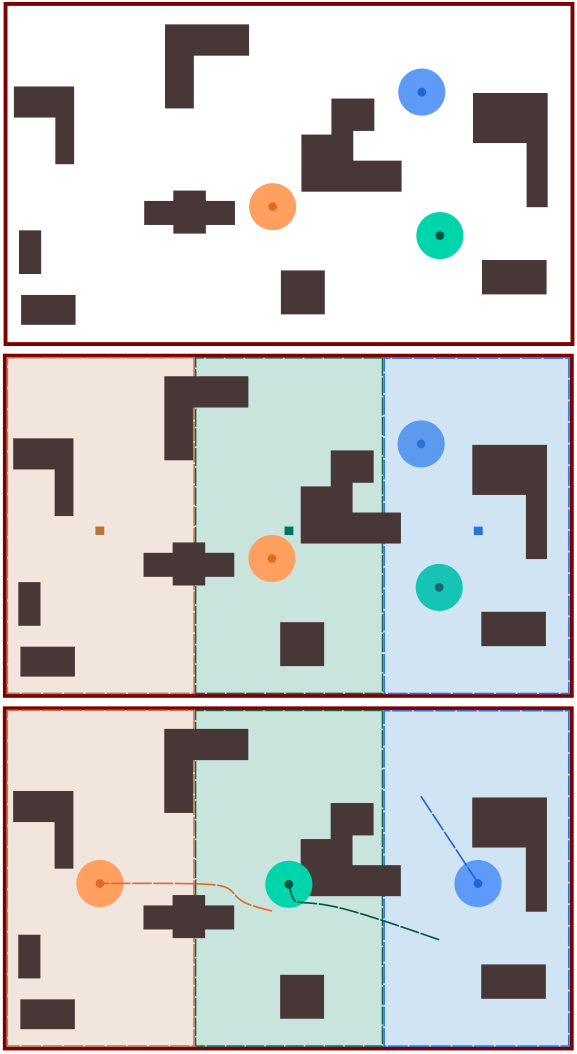}}
        ,(50,103)*{\text{\footnotesize \bf Rectangular}}
      \end{xy}
    \end{minipage}
    
    \caption{Each algorithm is depicted in three panels where: the top panel represents the start, center panel depicts a time midway through the algorithm's execution, and the bottom panel shows the ending state and paths taken. Paths over buildings are red. In the Voronoi and Rectangular methods, the regions are colored based on their corresponding robot.}
    \label{fig:algorithm_fig}
\end{figure}

Multi-agent coverage requires a team of agents to collectively perform a region-based mission, such as sweeping, where the mission objective considers the measure of the sub-region in which the task is performed. Depending on the mission, the coverage region may be defined as a subset of space, time, or \mbox{space $\times$ time}. 
Examples include: vacuuming a building, streaming scientific measurements from the field, and monitoring a secured area for intruders. 

In this paper we consider multi-agent coverage problems of surveillance and patrol in urban environments. We assume a team of identical UAV agents, such that each agent is equipped with a downward facing camera. The team is tasked with visually monitoring the subset of the environment that is outside of buildings at ground level.

Three dimensional Urban environments present  unique challenges for coverage using small UAVs. This is true even in the special case of ground level coverage.
Buildings are obstacles to navigation that agents must fly over or around.  Increasing an agent's altitude is assumed to have two effects on sensing ability. First, raising the height of the downward-facing senor cone increases the size of the sensor's ground-level footprint. Second, the increased footprint size causes loss of ground-level detail because larger patches of ground-level area are mapped to image pixels. 

We assume that due to loss of ground level detail the agents' camera sensors are not useful above a predefined maximum altitude. We also assume that this maximum altitude is below the height of the tallest building in the environment (otherwise the navigational constraints of buildings are trivially ignored by having all agents fly above all buildings all the time). We are interested in problems where agents must temporarily alter their paths to fly around or over buildings, yet where flying over a building prevents an agent's sensor from contributing to the monitoring task.

%
In {\it static multi-agent coverage} agents move to a set of mutually advantageous locations, and then remain at the same locations for the duration of the mission. In {\it  dynamic multi-agent coverage}, agents move continuously throughout the environment so that every point in the environment is intermittently observed. 
Many missions exist that are inherently static or dynamic in nature. 
Perimeter monitoring \cite{Gupta.etal.MRS19} requires that any point that {\it is} monitored must be monitored continuously (static coverage). Autonomous vacuuming \cite{Wong.etal.ACRA02} requires that every point in the environment be swept as frequently as resources allow (dynamic coverage). 
However, such an obvious choice may not exist for other coverage problems, and the relative applicability of a static or dynamic solution may depend on the problem instance. 

\extraDiscussion{Both static and dynamic coverage can be used to for urban monitoring and patrol. Their relative usefulness will depend on the relationship between: the total area that must be monitored, the number of agents available, and the sensor footprint of each agent.  
When the size of the environment is relatively large with respect to [team size $\times$ sensor footprint] we are forced to choose between monitoring a small subset of the environment continuously or a larger subset of the environment intermittently.}

\extraDiscussion{
While deterministic sweep algorithms are useful for many coverage applications (painting, vacuuming), they are less suited to adversarial surveillance tasks.  
}
Knowledge of how a deterministic path is generated can enable an adversary to evade detection. This problem can be mitigated by making paths (and thus revisit times) non-deterministic. 
However, a simple random walk is inefficient because the time to cover the environment is relatively large \cite{Feige.RSA95}.
{\it Ergodic sweep} balances randomness with coverage efficiency to achieve practically useful revisit properties \cite{a5}.

{\bf We experimentally evaluate six multi-agent coverage methods in simulation, including: three versions of Multi-agent ergodic coverage (differing in how obstacles are addressed), lawn-mower sweep, a voronoi-based static method, and a naive grid-based method. 
All six methods are experimentally evaluated with respect to four performance metrics, across four types of urban environments (two different building heights, and two different building densities), and for team sizes ranging from 2 to 25 agents.}

This paper is organised as follows: Section~\ref{sec:relatedWork} contains related work. Preliminaries, including nomenclature and a formal problem definition, appear in Section \ref{sec:preliminaries}. The algorithms we compare are described in Section \ref{sec:algorithms}. Experiments appear in section~\ref{sec:experiments}, and a discussion of result appears in Section~\ref{sec:results}. Section~\ref{sec:conclusions} concludes the paper.

\section{Related Work} \label{sec:relatedWork}

We assume a broad definition of the term `coverage,' and use it to indicate the set of problems including {\it coverage}, {\it surveillance}, and {\it environmental monitoring}. 
\extraDiscussion{Coverage methods are often divided into the two categories of dynamic coverage and static coverage.} 
In {static coverage} agents achieve stationary positions that enable the environment to be continuously monitored \cite{r19,r20,r21,r23}. In {dynamic coverage} agents move continuously through the environment so that all points are visited \cite{a1,a3,a4,a5,a6}. Depending on the scenario, points may be visited once (painting, vacuuming), or multiple times (patrol). 
\extraDiscussion{We now survey static coverage, dynamic coverage, and other related problems, respectively.}

\subsection{Dynamic Coverage}

{\it Lawn mower sweep} or {\it Boustrophedon Coverage} uses a simple back and forth motion. Assuming a sweep of positive radius, this can be used to cover any obstacle-free convex environment of finite area \cite{a1}. Polygonal obstacles can be addressed by partitioning the free space into a finite number of convex regions, and sweeping each region separately \cite{a3}. 
%
%
%
In  \cite{a4} a multi-agent team is divided into two groups of agents, one performing exploration and the other coverage. 
%
%
Most lawn mower sweep implementations assume a 2D environment such that regions surrounded by obstacles are topologically separated from the rest of the environment. In the 3D urban scenarios we consider
\extraDiscussion{we project a lawn mower sweeping path down onto the environment and then stitch together discontinuities in elevation.
In other words,}%
agents can adjust their elevation%
\extraDiscussion{along each lawn mower path }
to fly up and over buildings%
\extraDiscussion{. This enables agents }
to observe topologically separated ground-level regions.

Alternative methods of generating a deterministic coverage path include: Zelinski's algorithm, which orders the sweep based on the levels sets of a wave-front expansion \cite{Zelinsky.etal.ICRA93}, and ideas motivated by space filling curves \cite{Spires.Goldsmith.CR98}.

Ergodic sweep is applicable to adversarial coverage problems, such as patrol \cite{a5}, and uses nondetermisitc control laws to balance search efficiency and unpredictability (since, in an adversarial scenario, the practice of revisiting points on a set schedule can be exploited by the adversary).
Ergodic sweep control laws are often designed as a function of a target coverage distribution and the current coverage distribution.
%
%
Biasing coverage to obstacle-free  \cite{a5} reduces obstacle collisions, but does not eliminate collisions. 
A method for explicit obstacle avoidance within Ergodic coverage is presented in \cite{a6}, which combines ergodic coverage control laws with phisiocomemetic vector fields to ``repel'' agents away from obstacles. 
Three of the six algorithms that we compare are Ergodic coverage variants that, respectively: (1) ignore obstacles during coverage planning and then fly up and over them, (2) biases search away from obstacles and then fly up and over them if necessary, (3) avoid obstacle by flying around them without a change in elevation.

\subsection{Static Coverage}

Static coverage algorithms partition the search space into regions such that there is a mapping from robots to regions, and each robot can view its region from a stationary position. \extraDiscussion{Methods differ in how the space is divided, and how the robot-to-region mapping is calculated.}

Centralized methods \cite{r19,r20,r22} compute the search space division on a single agent or server, and then send the solution to all agents at runtime or {\it a priori}. 
A cellular decomposition of the environment followed by the calculation of a multi-robot spanning forest is used by \cite{r20}, and each robot is assigned one tree in the forest. 
%
%
%
A segmentation  technique  is used in \cite{r22}, where each robot is assigned a unique segment.
Work in \cite{r9} partitions the environment using the weighted K-Means clustering algorithm. 
Centralization has the disadvantage of introducing single points of failure.

In contrast, decentralized and/or distributed process do not have a single point-of-failure. In many strategies robots start at their initial locations and then tend toward a multi-robot configuration with desirable coverage properties over time.  
Iterative Voronoi-based approaches  achieve such a distributed control strategy 
\cite{%
r4,%
r7,%
r14
}. 
\extraDiscussion{At the start of the mission, robot positions are used as the generator points of a Voronoi decomposition of the search space, then each robot moves a small distance toward the center of the Voronoi cell that its generated. The process is then repeated during each time step, and with the result that agents tend to space themselves evenly throughout the environment.}
One advantage of this method is that each agent need only communicate with its Voronoi neighbors. 
The basic Voronoi method is one of the six methods we compare in this paper.

\subsection{Other Problems Related to Coverage}

Information gathering \cite{Julian.etal.IJRR12,Khan.etal.CNS15} is closely related to coverage, the main difference being that the amount of information that can be gathered at a particular location is non-uniform and changes as a function of each sensor measurement.

In target search \cite{Koopman.OR56,Chung.etal.AR11,Waharte.Trigoni.est10,Nayak.etal.RAL19} 
the goal is to locate one or more stationary or moving targets.
%
Target search problem variants that share similarities with the coverage problem include probability based 
\cite{Chung.Burdick.TRO12,%
Dames.Kumar.TASE15}, 
information based 
\cite{Dames.etal.CDC12,%
Hollinger.etal.TRO15,%
Berger.Happe.CEC10}, and
game theoretic formulations \cite{Otte.etal.AR17}. 

%

Exploration is similar to coverage in that both problems are concerned with visiting all points in the environment \cite{Burgard.etal.ICRA00}. However, coverage problems often require repeated visits to each location, while exploration is completed once each location  has been visited once.

In previous work \cite{Arul.etal.RAL19} we present an  collision avoidance algorithm for a swarm of UAVs performing an urban coverage task. The method in \cite{Arul.etal.RAL19} focuses on agent-to-agent and agent-to-obstacle local collision avoidance. The method in \cite{Arul.etal.RAL19} can be used as a post-processing step for all of the algorithms considered in the current paper.

\section{Preliminaries} \label{sec:preliminaries}


In Section~\ref{sec:nomenclature} we define our nomenclature and assumptions, and in Section \ref{sec:problemStatement} we formally define the problems of static and dynamic multi-agent urban coverage.

\subsection{Nomenclature and Assumptions} \label{sec:nomenclature}

A team of $\numRobots$ robots is the set ${\robotTeam = \{\robot{1}, \ldots, \robot{\numRobots}\}}$, where $\robot{\robotIndex}$ denotes the $\robotIndex$-th robot. 
The 3D workspace in which the robots operate is $\spacespacetd{}$. 
The {\it obstacle space} ${\spacespacetd{obs} \subset \spacespacetd{}}$ is the subset of $\spacespacetd{}$ containing obstacles, while
the {\it free space} ${\spacespacetd{free}  \subset \spacespacetd{}}$ is the subset of $\spacespacetd{}$ that does not contain obstacles.  ${\spacespacetd{obs} \cap \spacespacetd{free} = \emptyset}$ and ${\spacespacetd{obs} \cup \spacespacetd{free} = \spacespacetd{}}$.
We assume that the ground plane of the workspace can be approximated by 2D Euclidean space $\reals^2$. 
Let the projections of $\spacespacetd{}$ and $\spacespacetd{obs}$ directly down onto $\reals^2$ be denoted $\spacespace{}$ and $\spacespace{obs}$, respectively.
The free space at ground level is defined ${\spacespace{free} = \spacespace{} \setminus \spacespace{obs}}$ such that ${\spacespace{obs} \cap \spacespace{free} = \emptyset}$ and ${\spacespace{obs} \cup \spacespace{free} = \spacespace{}}$.
We assume an urban environment such that $\spacespacetd{obs}$ contains buildings. The area of desired coverage is defined to be all ground-level terrain (and not the sides or tops of buildings), and denoted $\spacespace{search} \equiv \spacespace{free}$.

Each robot  $\robot{\robotIndex}$ is assumed to be a point, and moves along a path in $\spacespacetd{free}$ and observes $\spacespace{free}$ using a downward facing camera sensors. 
We assume a continuous time model starting at $\timeVariable=0$. The interval of time from the beginning of the mission until ${\timeVariable=\timeVariableMax}$ is $[0, \timeVariableMax]$.
A valid robot path from time ${\timeVariable=0}$ to time ${\timeVariable=\timeVariableMax}$ is a continuous function ${\motionPath{\robotIndex} : [0, \timeVariableMax] \,\rightarrow \,\spacespacetd{free}}$.  
For problems involving continuous or continual coverage, we often make the tacit assumption that ${\timeVariableMax \rightarrow \infty}$. In other words, that $\motionPath{\robotIndex}$ continues to be well defined as search time increases without bound. (This contrasts with the typical path planning problem, in which paths have well defined start {\it and} end points and can thus be represented as maps from the unit interval $[0, 1]$).

For convince, we abuse our notation and (also) let path $\motionPath{\robotIndex}$  denote the geometric set of points along the curve in $\spacespacetd{free}$ that is traced out by robot $\robot{\robotIndex}$ over time, such that ${\motionPath{\robotIndex} \subset \spacespacetd{free}}$.
The multipath $\pathset$ is the set of all robots' paths. $\pathset$ is notationally overloaded in the same way as $\motionPath{}$. 
As a function, ${\pathset : [0, \timeVariableMax] \,\rightarrow \, (\spacespacetd{free})^\numRobots}$, where 
$\spacespacetd{free,\robotIndex}$ is a copy of the 3-dimensional free space associated with robot $\robot{\robotIndex}$
and
${(\spacespacetd{free})^\numRobots = \spacespacetd{free,1} \times \ldots  \times \spacespacetd{free,\numRobots}}$
is the product space of all robot's free spaces. In the geometric sense ${\pathset = \bigcup_{\robotIndex = 1}^{\numRobots} \{\robot{\robotIndex}\}}$.
The space containing all multipaths is denoted $\multiPathSpace = \bigcup \{\pathset \}$.

We assume robots have identical downward facing camera sensors such that when $\robot{\robotIndex}$ flies at altitude $\altitude$ the projection of $\robot{\robotIndex}$'s field-of-view down onto $\spacespace{search}$ is a disc $\disc_{\altitude}$ of radius $f(\altitude)$.
We assume there exists an optimal altitude $\optimalAltitude$ for the camera sensor to be used. For example, a sensor will be used at the highest altitude for which it remains a reliable sensor (increasing altitude increases $\disc_{\altitude}$, which is advantageous, but may degrade sensor reliability, which is disadvantageous). We drop the subscript when the sensor is used at the optimal altitude, $\disc = \disc_{\optimalAltitude}$.
Thus, while observing $\spacespace{free}$ agents are assumed to fly at altitude $\optimalAltitude$  in $\spacespacetd{free}$. Agents may increase their altitude ${\altitude > \optimalAltitude}$  to fly over obstacles. We assume sensors cannot be used to observe  $\spacespace{free}$ (the ground-plane) during maneuver at increased altitude ${\altitude > \optimalAltitude}$' in particular, during maneuvers up and over buildings.
Let ${\motionPathObserving{\robotIndex} \subset \motionPath{\robotIndex}}$ be the subset of $\motionPath{\robotIndex}$ containing all points at which robot $\robot{\robotIndex}$'s camera sensor is functional, i.e., for which $\robot{\robotIndex}$ is at altitude ${\altitude = \optimalAltitude}$. Similarily, let $\pathsetObserving = \bigcup_{\robotIndex = 1}^{\numRobots} \{\motionPathObserving{\robotIndex} \}$. The area swept by the team is then:
\begin{equation}\label{eq:sweptArea}
    \spacespace{swept} = \left( \pathsetObserving  \minkowskiSum  \disc \right) \cap \spacespace{search}
\end{equation}
where `$\minkowskiSum$' denotes the Minkowski sum. The intersection with $\spacespace{search}$ is included in Equation~\ref{eq:sweptArea} so that ${\spacespace{swept} \subset \spacespace{search}}$ by construction.

Let $\multiPathObservingSpace = \bigcup \{ \pathsetObserving  \}$ be the space containing all $\pathsetObserving$.
Let $\indicatorFunction$ be an indicator function that returns $1$ or $0$ based on whether or not, at time $\timeVariable$, a point ${\spacePoint \in \spacespace{search}}$ is observed by at least one robot in the team.  
${\indicatorFunction \,:\, \spacespace{search} \times [0, \timeVariableMax] \times \multiPathObservingSpace \rightarrow \{1, 0\}}$.
and so ${{\indicatorFunction(\spacePoint, \timeVariable, \pathsetObserving) \mapsto 1} \iff \spacePoint \in \pathsetObserving(\timeVariable) \minkowskiSum \disc
}$ and
${
{\indicatorFunction(\spacePoint, \timeVariable, \pathsetObserving) \mapsto 0} \iff \spacePoint \not\in \pathsetObserving(\timeVariable) \minkowskiSum \disc
}$.

Given a particular $\pathsetObserving$, the function $\totalVisitDuration$ is a map from the search space $\spacespace{search}$ to the time-duration domain, ${\motionPath{\robotIndex} : \spacespace{search}  \,\rightarrow \, [0, \timeVariableMax]}$ that measures the cumulative time 
that 
point  ${\spacePoint \in  \spacespace{search}}$ is observed by at least one robot,
${
\totalVisitDuration(\spacePoint) = \int_0^{\timeVariableMax}  \indicatorFunction(\spacePoint, \timeVariable, \pathsetObserving) \, d\timeVariable
}$,
where the integral is Lebesgue. Dynamic coverage algorithms can be characterized by a requirement that points in $\spacespace{search}$ be visited infinity often as ${\timeVariableMax \rightarrow \infty}$. Given 
continuous vehicles paths, this is 
$
\lim_{\timeVariableMax \rightarrow \infty} \totalVisitDuration(\spacePoint) = \infty
$ for all ${\spacePoint \in \spacespace{search}}$.

In static search we seek to maximize the number of points continually visible to at least one robot. For search times $\timeVariableMax$ large enough that the complicating effects of the startup phase can be ignored (in which agents move to their optimal static coverage locations), this is equivalent to finding  
${
{\argmax_{\pathset \in \multiPathSpace} \frac{1}{\timeVariableMax} \int_{\spacePoint \in \spacespace{search}} \totalVisitDuration(\spacePoint)  \, d\spacespace{search}}
}$
i.e., by calculating the static robot positions $\{\spacePoint_1, \ldots \spacePoint_\numRobots\}$ that maximize the Lebesgue measure of $\spacespace{search} \cap \bigcup_{\robotIndex = 1}^{\numRobots} \spacePoint_\robotIndex \minkowskiSum \disc_\robotIndex$.

\subsection{Problem Statements} \label{sec:problemStatement}

We consider algorithms designed to solve {\it static} or {\it dynamic} variants of the coverage problem. 

\noindent \textbf{Problem 1}, The \textit{dynamic multi-robot urban coverage problem}: Given a team of $\numRobots$ robots ${\robotTeam = \{\robot{1}, \ldots, \robot{\numRobots}\}}$, an urban environment ${\spacespacetd{}}$ with obstacle space and free space such that ${\spacespacetd{obs} \cap \spacespacetd{free} = \emptyset}$ and ${\spacespacetd{obs} \cup \spacespacetd{free} = \spacespacetd{}}$, and with ground-level search space $\spacespace{search} = \spacespace{free}$, determine a multipath ${\pathset = \{\robot{1}, \ldots, \robot{\numRobots}\}}$ such that $\spacespace{swept} = \left( \pathsetObserving  \minkowskiSum  \disc \right) \cap \spacespace{search}$ and where,  for all ${\spacePoint \in \spacespace{search}}$, ${\lim_{\timeVariableMax \rightarrow \infty} \totalVisitDuration(\spacePoint) = \infty}$.

\noindent \textbf{Problem 2}, The \textit{static multi-robot urban coverage problem}: Given a team of $\numRobots$ robots ${\robotTeam = \{\robot{1}, \ldots, \robot{\numRobots}\}}$, an urban environment ${\spacespacetd{}}$ with obstacle space and free space such that ${\spacespacetd{obs} \cap \spacespacetd{free} = \emptyset}$ and ${\spacespacetd{obs} \cup \spacespacetd{free} = \spacespacetd{}}$, and with ground-level search space $\spacespace{search} = \spacespace{free}$, find 
${
\argmax_{\pathset \in \multiPathSpace} \frac{1}{\timeVariableMax} \int_{\spacePoint \in \spacespace{search}} \totalVisitDuration(\spacePoint)  \, d\spacespace{search}
}$.

\section{Algorithms} \label{sec:algorithms}

\newcommand{\eat}[1]{}

We now describe the six multi-agent methods we compare.
Lawn mower, ergodic, Voronoi, and grid-based methods appear in Sections \ref{sec:lawnMower}, \ref{sec:ergodic}, \ref{sec:voronoi}, and \ref{sec:grid}, respectively.

\subsection{Multi-Agent Urban Lawn Mower Coverage} \label{sec:lawnMower}

Single agent lawn mower sweep in a 2-dimensional environment is known as boustrophedon coverage. Our multi-agent implementation is presented in Algorithm~\ref{alg:MultiLawnMower}, and uses the single agent boustrophedon algorithm as a subroutine. We start by finding a single agent boustrophedon coverage cycle of an obstacle free environment $\spacespace{vacant}$, where $\spacespace{vacant}$ has the same footprint as $\spacespace{}$ \mbox{(line \ref{alg:MultiLawnMower:a})}. Next, we repair this solution to avoid obstacles in $\spacespacetd{}$ using the subroutine $\flyOverBuildings{\motionPath{},\spacespacetd{free},\spacespacetd{obs}}$ on line~\ref{alg:MultiLawnMower:b}. The effect of this subroutine is to calculate a building avoiding path $\motionPathObserving{}$ by replace each obstacle intersecting segment of the original path $\motionPath{}$   with a different segment along which the agent increase its elevation above the building, flies over it, and then descends back to the optimal sweep elevation. Given a single agent cycle $\motionPathObserving{1}$ of length $\pathLength$, the cycles of the other agents are calculated by rotating the starting positions $\frac{\robotIndex}{\numRobots}$ of the length around the cycle for agent $\robotIndex$ (lines \ref{alg:MultiLawnMower:c}\ref{alg:MultiLawnMower:e}). The obstacle avoiding multi-path of the team is given by the $\numRobots$-tupal containing the single agent paths (line \ref{alg:MultiLawnMower:f}).

\extraDiscussion{At run-time, each agent follows its own version of the cycle. Agents that do not start the mission at the beginning of their cycles head directly to the nearest point along the cycle at which they will be on schedule. Each agent $\robotIndex$ repeatedly moves around its cycle $\motionPathObserving{1}$.}

\newcommand{\algScaleSize}{.83}

\begin{figure*}[t!]

\scalebox{\algScaleSize}{%
\begin{minipage}{.4\textwidth}

\begin{algorithm}[H]
  \begin{algorithmic}[1]
    \algorithmTextSize
    \caption{\mbox{Multi-Agent Lawn Mower}} \label{alg:MultiLawnMower}
    \STATE $\motionPath{} \gets \boustrophedon{\spacespace{vacant}}$\;   \label{alg:MultiLawnMower:a}
    \STATE $\motionPathObserving{} \gets \flyOverBuildings{\motionPath{},\spacespacetd{free},\spacespacetd{obs}}$\;  \label{alg:MultiLawnMower:b}
    \STATE $\pathLength \gets \|\motionPathObserving{1}\|$\;  \label{alg:MultiLawnMower:c}
    \FOR{$\robotIndex \gets 1 \ldots \numRobots$ \label{alg:MultiLawnMower:d} }
        \STATE $\motionPathObserving{\robotIndex} \gets  \rotateCycle{ \motionPathObserving{}, \frac{\robotIndex}{\numRobots}\pathLength }$\;   \label{alg:MultiLawnMower:e}
    \ENDFOR
    \STATE $\pathsetObserving \gets (\motionPathObserving{1}, \ldots, \motionPathObserving{\numRobots})$\;   \label{alg:MultiLawnMower:f}
  \end{algorithmic}
\end{algorithm}

\vspace{\algRaiseSize}


\begin{algorithm}[H]
  \begin{algorithmic}[1]
    \algorithmTextSize
    \caption{$\singleAgentErgodic{\pdfFunction}$}\label{alg:saergodic}
    \STATE $M_k, C_k \gets \initializeErgodicParams(\pdfFunction)$ \;
    \STATE $\motionPath{}  \gets \emptyset$ \;
    \FOR{$\timeVariable = 1, \ldots \timeVariableMax$}
        \STATE $\motionPath{[0, \timeVariable]} \gets \motionPath{[0, \timeVariable-1]} \concat \calculateNextStep(\pdfFunction, M_k, C_k)$ $\!\!\!\!\!$\;  \label{saergodic:c}
        \STATE $C_k \gets \updateCurrentDist(C_k, \motionPath{[\timeVariable]})$ \;
    \ENDFOR
  \end{algorithmic}
\end{algorithm}

\vspace{\algRaiseSize}


\begin{algorithm}[H]
  \begin{algorithmic}[1]
    \algorithmTextSize
    \caption{$\singleAgentErgodicObstacles{\pdfFunctionFree,\spacespace{obs}}$} \label{alg:saergodicobst}
    \STATE $M_k, C_k \gets \initializeErgodicParams(\pdfFunctionFree)$ \;
    \STATE $\motionPath{[0, \timeVariable]}  \gets \emptyset$ \;
    \FOR{$\timeVariable = 1, \ldots \timeVariableMax$}
        \STATE $\motionPath{[\timeVariable-1, \timeVariable]} \gets \calculateNextStep(\pdfFunctionFree, M_k, C_k)$ \;
        \STATE $\motionPath{[0, \timeVariable]} \gets \calculateVectorField(\motionPath{[\timeVariable-1, \timeVariable]}, \spacespace{obs})$ $\!\!\!\!\!$\; \label{saergodicobst:d}
        \STATE $C_k \gets \updateCurrentDist(C_k, \motionPath{[\timeVariable]})$ \;
    \ENDFOR
  \end{algorithmic}
\end{algorithm}

\end{minipage}}
\scalebox{\algScaleSize}{%
\begin{minipage}{.4\textwidth}


\begin{algorithm}[H]
  \begin{algorithmic}[1]
    \algorithmTextSize
    \caption{\mbox{Multi-Agent Ergodic}} \label{alg:NaieveErgodic}
    \FOR{$\robotIndex \gets 1 \ldots \numRobots$ \label{alg:NaieveErgodic:1} }
      \STATE $\motionPath{\robotIndex} \gets \singleAgentErgodic{\pdfFunctionVacant}$\;   \label{alg:NaieveErgodic:b}
      \STATE $\motionPathObserving{\robotIndex} \gets \flyOverBuildings{\motionPath{\robotIndex},\spacespacetd{free},\spacespacetd{obs}}$\;  \label{alg:NaieveErgodic:c}
    \ENDFOR
    \STATE $\pathsetObserving \gets (\motionPathObserving{1}, \ldots, \motionPathObserving{\numRobots})$\;   \label{alg:NaieveErgodic:d}
  \end{algorithmic}
\end{algorithm}


\vspace{\algRaiseSizeB}

\begin{algorithm}[H]
  \begin{algorithmic}[1]
    \algorithmTextSize
    \caption{\mbox{Biased Multi-Agent Ergodic}} \label{alg:BiasedErgodic}
    \FOR{$\robotIndex \gets 1 \ldots \numRobots$ \label{alg:BiasedErgodic:1} }
      \STATE $\motionPath{\robotIndex} \gets \singleAgentErgodic{\pdfFunctionFree}$ \;   \label{alg:BiasedErgodic:b}
      \STATE $\motionPathObserving{\robotIndex} \gets \flyOverBuildings{\motionPath{\robotIndex},\spacespacetd{free},\spacespacetd{obs}}$ \;  \label{alg:BiasedErgodic:c}
    \ENDFOR
    \STATE $\pathsetObserving \gets (\motionPathObserving{1}, \ldots, \motionPathObserving{\numRobots})$ \;   \label{alg:BiasedErgodic:d}
    \end{algorithmic}
\end{algorithm}

\vspace{\algRaiseSizeB}

\begin{algorithm}[H]
  \begin{algorithmic}[1]
    \algorithmTextSize
    \caption{\newline{Obstacle Avoiding Multi-Agent Ergodic}} \label{alg:AvoidingErgodic}
    \FOR{$\robotIndex \gets 1 \ldots \numRobots$ \label{alg:AvoidingErgodic:1} }
      \STATE $\motionPath{\robotIndex} \gets \singleAgentErgodicObstacles{\pdfFunctionFree,\spacespace{obs}}$ \;   \label{alg:AvoidingErgodic:b}
    \ENDFOR
    \STATE $\pathsetObserving \gets \pathset \gets (\motionPath{1}, \ldots, \motionPath{\numRobots})$ \;   \label{alg:AvoidingErgodic:d}
    \end{algorithmic}
\end{algorithm}

\end{minipage}}
\scalebox{\algScaleSize}{%
\begin{minipage}{.385\textwidth}


\begin{algorithm}[H]
  \begin{algorithmic}[1]
    \algorithmTextSize
    \caption{\mbox{Multi-Agent Voronoi Cover}} \label{alg:alvoronoi}
        \STATE${\teamSpacePoint_0 \gets \projectionOnto(\robotTeam , \spacespace{})}$\;
        \STATE$\{\motionPath{0}, \ldots,  \motionPath{\numRobots}  \} \gets \{\spacePoint_{0,1}, \ldots, \spacePoint_{0,\numRobots}\} \gets \teamSpacePoint_0  $\;
        \FOR{$\timeVariable \gets 1, \ldots, \timeVariableMax $   \label{alvoronoi:z}}
            \STATE ${\{\voronoiCell_1, \ldots, \voronoiCell_\numRobots \} \gets \voronoiPartition(\teamSpacePoint_{\timeVariable-1}, \spacespace{})}\!\!\!\!\!\!\!\!\!\!$\; \label{alvoronoi:a}
            \FOR{$\robotIndex \gets 1, \ldots, \numRobots $}
                \STATE $\spacePoint_{\timeVariable, \robotIndex} \gets \centroid(\voronoiCell_\robotIndex )$\;
                \STATE $\motionPath{\robotIndex} = \motionPath{\robotIndex} \concat \spacePoint_{\timeVariable, \robotIndex}$ \; \label{alvoronoi:c}
            \ENDFOR
            \STATE ${\teamSpacePoint_{\timeVariable} \gets \{\spacePoint_{\timeVariable,1}, \ldots, \spacePoint_{\timeVariable,\numRobots}\}}$\; \label{alvoronoi:d}
        \ENDFOR
        \FOR{$\robotIndex \gets 1, \ldots, \numRobots $}
            \STATE $\motionPathObserving{\robotIndex} \gets \flyOverBuildings{\motionPath{\robotIndex},\spacespacetd{free},\spacespacetd{obs}}$ \;
        \ENDFOR
        \STATE $\pathsetObserving \gets (\motionPathObserving{0}, \ldots,  \motionPathObserving{\numRobots})$ \;
        
    \end{algorithmic}
\end{algorithm}

\vspace{\algRaiseSizeC}

\begin{algorithm}[H]
  \begin{algorithmic}[1]
    \algorithmTextSize
    \caption{\mbox{Multi-Agent Grid Cover}} \label{alg:grid}
        \STATE${\teamSpacePoint_0 \gets \projectionOnto(\robotTeam , \spacespace{})}$\;
        \STATE$\{\motionPath{0}, \ldots,  \motionPath{\numRobots}  \} \gets \{\spacePoint_{0,1}, \ldots, \spacePoint_{0,\numRobots}\} \gets \teamSpacePoint_0  $\;
        \STATE${\{\voronoiCell_1, \ldots, \voronoiCell_\numRobots \} \gets \gridPartition(\teamSpacePoint_{0}, \spacespace{})}\!\!\!\!\!\!\!\!\!\!$\;
        \FOR{$\robotIndex \gets 1, \ldots, \numRobots $}
            \STATE$\spacePoint_{1, \robotIndex} \gets \centroid(\voronoiCell_\robotIndex )$\;
            \STATE$\motionPath{\robotIndex} = \motionPath{\robotIndex} \concat \spacePoint_{1, \robotIndex}$\;
            \STATE$\motionPathObserving{\robotIndex} \gets \flyOverBuildings{\motionPath{\robotIndex},\spacespacetd{free},\spacespacetd{obs}}$ \;
        \ENDFOR
        \STATE$\pathsetObserving \gets (\motionPathObserving{0}, \ldots,  \motionPathObserving{\numRobots})$ \;
  \end{algorithmic}
\end{algorithm}

\end{minipage}}
\end{figure*}

\subsection{Multi-Agent Urban Ergodic Coverage} \label{sec:ergodic}

Non-determinism is useful in scenarios in which an adversary attempts to avoid detection.
%
In Ergodic coverage the agent/team follows non-deterministic trajectories such that the relative time spent in each non-zero measure region of the environment can be prescribed by a user.
The desired properties only hold almost surely in the limit as time approaches infinity. Thus, practical performance can be expected to improve with mission duration.
%

Ergodic coverage is most easily described---and implemented---as an evolutionary process that generates a path. 
%
%
The subroutine  $\singleAgentErgodic{\pdfFunction}$, described in Algorithm~\ref{alg:saergodic}, computes the ergodic path for a single agent, assuming a user defined coverage distribution ${\pdfFunction(\spacePoint)}$ over $\spacespace{}$ is desired.  The `$\concat$' symbol denotes path concatenation. 

The subroutine $\calculateNextStep(\pdfFunction, M_k, C_k)$ (on line \ref{saergodic:c} of Algorithm \ref{alg:saergodic}) implements 
the control laws of ergodic coverage for a single agent without explicit obstacle avoidance. This is calculated
${
    B_j(t) = \sum_{R} \Lambda_{k} S_{k} \nabla f_k(x_j(t))
}$
which is further normalized and constrained by velocity $u_{max}$
${
    u_j(t) = - u_{max} \frac{B_j(t)}{\|B_j(t)\|_2}
}$,
Where $\Lambda_{k}$ is constant and $S_k(t)$ is difference between the current distribution and target distribution, given by $S_k(t) := C_k(t) - M_k(t)$. $\nabla f_k(x_j(t))$ is gradient of the Fourier basis function which is given by
$$
\nabla f_k(x_j(t)) = \frac{1}{h_k} \left[ 
\begin{array}{c}
    -k_1 \sin(k_1x_1)\cos(k_2x_2) \\
    -k_2 \cos(k_1x_1)\sin(k_2x_2) 
\end{array}\right]
$$
%
%
and the Fourier basis function is given by 
$
f_k(x) = \frac{1}{h_k} \cos(k_1x_1)\cos(k_2x_2).
$
The Fourier coefficient $C_k(t)$ is calculated
$
C_k(t) =  \sum_{j=1}^{N} \int_{0}^{t} f_k(x_j(\tau))d\tau
$
and Fourier coefficient of target distribution is
$M_k(t) := Nt\fourcoeffu$, where
$k_1 = \frac{K_1\pi}{L_1}$, $k_2 = \frac{K_2\pi}{L_2}$, and
$\fourcoeffu = \langle \mu, f_k \rangle$, where $\langle \cdot,\cdot\rangle$ is an inner product.
$h_k = \left( \int_{0}^{L_1} \int_{0}^{L_2} \cos^2(k_1x_1) \cos^2(k_2x_2) \right)^{1/2}$.

The subroutine  $\singleAgentErgodicObstacles{\spacespace{},\spacespace{obs}}$, described in Algorithm~\ref{alg:saergodicobst}, computes an ergodic path for a single agent that explicitly avoids obstacles by using an obstacle repulsive feedback law (line \ref{saergodicobst:d}), which is calculated:
This feedback law implementing obstacle avoidance would be governed by
${
V^*_j(t) := - \alpha V_j(t) + (1 - \alpha) F^o_j(r_j)
}$
where ${V_j(t) := - \frac{B_j(t)}{\|B_j(t)\|_2}}$ and $F^o_j(r_j)$ is a repulsive vector field. This is further normalized and constrained by velocity $u_{max}$ as 
${
u_j(t) = - u_{max} \frac{V_j(t)}{\|V_j(t)\|_2}
}$.
The parameter $\alpha \in [0,1]$  is a bump factor taking value 1 if the robot is far from the obstacle and reducing to 0 when approaching an obstacle (in the vector field). There are many ways one can define $\alpha$, in this paper we have defined it linearly. 

Let the probability distribution functions $\pdfFunctionVacant(\spacePoint)$ and $\pdfFunctionFree(\spacePoint)$ respectively define uniform random distribution over the entire space (ignoring obstacles) and over the free space (biasing movement away from obstacles). 
$$\pdfFunctionVacant(\spacePoint) = 
\begin{cases}  
  \frac{1}{\| \spacespace{} \|} & \text{if } \spacePoint \in  \spacespace{} \\
  0                         &  \text{if }  \spacePoint \not\in \spacespace{}
\end{cases}$$
$$\pdfFunctionFree(\spacePoint) = 
\begin{cases}  
  \frac{1}{\| \spacespace{free} \|} & \text{if } \spacePoint \in  \spacespace{free} \\
  0                         &  \text{if }  \spacePoint \not\in \spacespace{free}
\end{cases}$$

We compare three versions of multi-agent ergodic sweep for that differ based on how urban obstacles are handled:
\begin{enumerate}
\item \textbf{Multi-Agent Urban Ergodic Sweep}
%
is a simple repairing strategy that calculates a path assuming no obstacles exists (Algorithm~\ref{alg:NaieveErgodic}). This method handles obstacle avoidance in the same way as the multi-agent lawn mower sweep algorithm described in the previous section.  $\motionPathObserving{\robotIndex}$ is created from  $\motionPath{\robotIndex}$ by having agent $\robotIndex$ rise in elevation to fly over a building and descend back to the sweep altitude afterward (Algorithm~\ref{alg:NaieveErgodic}, line \ref{alg:NaieveErgodic:c}).

\item \textbf{Multi-Agent Urban Biased Ergodic Sweep}
%
is similar, 
except that we define the area of desired uniform coverage to be $\spacespace{free}$ instead of $\spacespace{vacant}$ (Algorithm~\ref{alg:BiasedErgodic}, line \ref{alg:NaieveErgodic:b}). This biases $\motionPath{\robotIndex}$ away from obstacles, but does not eliminate the need to occasionally fly up and over a buildings. 
$\motionPathObserving{\robotIndex}$ is created from  $\motionPath{\robotIndex}$ by having agent $\robotIndex$ fly over buildings when necessary (Algorithm~\ref{alg:BiasedErgodic} line \ref{alg:NaieveErgodic:c}). 

\item \textbf{Multi-Agent Obstacle Avoiding Urban Ergodic Sweep}
%
forces each $\motionPath{\robotIndex}$ to avoid obstacles using the vector field algorithm (Algorithm~\ref{alg:AvoidingErgodic}). In other words, by using $\singleAgentErgodicObstacles{\pdfFunctionFree,\spacespace{obs}}$ to prevent $\motionPath{\robotIndex}$ from intersecting $\spacespace{obs}$. In our experiments, all agents fly at the optimal sweep elevation such that  isolated internal courtyards are not visited. 
\end{enumerate}


\subsection{Voronoi Urban Coverage Algorithm} \label{sec:voronoi}

Given a set of $\numRobots$ generating points (we use the set $\teamSpacePoint = \{\spacePoint_{1}, \ldots, \spacePoint_{\numRobots}\}$ containing robot projections onto $\spacespace{}$) a Voronoi space partitioning of $\spacespace{}$ creates mutually disjoint set of cells by assigning each point ${\spacePoint \in \spacespace{}}$  to a cell $\voronoiCell_\robotIndex$ associated with the closest generating point (Algorithm \ref{alg:alvoronoi}, line \ref{alvoronoi:a}). 
${
\voronoiCell_\robotIndex =  \{\spacePoint \, | \,  \| \spacePoint - \spacePoint_{\robotIndex} \| <  \| \spacePoint - \spacePoint_{ \otherRobotIndex \neq \robotIndex}\|  \}
}$
The voroni partitioning of $\spacespace{}$, i.e., instead of all of $\reals^2$, is found by placing reflected versions of points across each boundary of $\spacespace{}$, and then truncating the resulting extended Voronoi diagram to the footprint of $\spacespace{}$. When $\spacespace{}$ is a rectangle this requires $5n$ points.

Voronoi coverage methods work by having each robot move a small distance toward the centroid of its current voroni cell, recalcuate new voronoii cells (Algorithm \ref{alg:alvoronoi}, lines \ref{alvoronoi:a}-\ref{alvoronoi:c}), and then repeat (lines \ref{alvoronoi:z}-\ref{alvoronoi:d}).
Over time, this causes robots to greedily space themselves away from their neighbors.

\subsection{Grid-Based Urban Coverage Algorithm} \label{sec:grid}

We also implement a naive grid-based algorithm for static coverage. This method  uses rectangles, instead of Voronoi regions, to divide the space among agents (Algorithm~\ref{alg:grid}).
%
%
While the Voronoi region algorithm is the result of a distributed control process, this rectangular division of the space can be calculated {\it a priori}. Each agent simply moves to the center of its assigned rectangular region. 
%

\section{Experiments} \label{sec:experiments}

In this section we compare the six multi-agent coverage methods (see Tables~\ref{table:algs} and \ref{table:metrics}) across a variety of environments (see Table \ref{table:environemnts}), for teams of size 1 to 25 agents. Environments differ based on building height, building density, and footprint size. Each combination of Environment, Algorithm, and team size is repeated over three random trials.

\begin{table}[t!]
\begin{minipage}[t]{3.43cm}
\centering
\caption{Algorithms} \label{table:algs} 
\resizebox{3.40cm}{!}{%
\begin{tabular}{| l | l |}
\hline
 & Name of Algorithm\\
\hline
1 & Lawnmower\\
2 & Ergodic\\
3 & Biased Ergodic\\
4 & Obstacle Avoiding Ergodic\\
5  & Voronoi \\
6 & Rectangular \\
\hline
\end{tabular}
}
\end{minipage}
\begin{minipage}[t]{5.15cm}
\centering
\caption{\mbox{Evaluation Metrics}} \label{table:metrics}
\resizebox{5.10cm}{!}{%
\begin{tabular}{| l | l |}
\hline
 & Name of Metric\\
\hline
1 & Coverage Area (percentage of total area)\\
2 & Number of Visits \\
3 & Mean Duration Between Visits to Each Points\\
4 & Mean Time Spent Near Each Point\\
\hline
\end{tabular}
}
\end{minipage}

\end{table}

\begin{table}[t!]
\centering

\caption{Urban Environments in Experiments} \label{table:environemnts}

\begin{tabular}{|l|c|c|c|c|}
\hline
& Height & Density & Dimensions & Buildings\\
\hline
1 & Tall & High & 50.96 $\times$ 39.33 $\times$ 29.50 & 27\\
2 & Tall & Low & 56.25 $\times$ 53.03 $\times$ 14.25 & 16\\
3 & Short & High & 64.26 $\times$ 53.80 $\times$ 12.50 & 79\\
4 & Short & Low & 96.67 $\times$ 62.92 $\times$ 7.2 & 23\\
5 & Mixed & Mixed & 147 $\times$ 59 $\times$  &\\
\hline
\end{tabular}

\end{table}


Experiments are run in the Ubuntu Linux operating system using Robot Operating System (ROS), Gazebo, and the pix4 control package. UAVs measure 1 X 1 X 0.3 meters. 



We empirically evaluate performance with respect to four metrics (described below), by randomly sampling a large number of points in the environment and tracking statistics in a disc surrounding each point.
Statistical results for a particular trial are obtained by integrating over these points.
%

\begin{enumerate}
\item \textbf{Percent Coverage}: 
%
The percentage of the map (point regions) that has been swept at least once. 
%

\item \textbf{Visits Count}.
The total number of visits a point's region has been visited. 
%

\item \textbf{Revisit Time}.
The time duration between successive visits to a point's region.
%

\item \textbf{Time spent}.
The cumulative time that any agent is within a point's region. We unpause the counter when any agent enters  point's region and pause the counter when the agent leaves that region. 
%
\end{enumerate}

Agent starting locations are chosen randomly for all methods except for Lawn Mower Sweep. For Lawn Mower Sweep, the initial coordinates of the agents are randomly chosen along the lawn mower path (this causes the lawnmower algorithm to have a slight start-up advantage%
\extraDiscussion{over other methods because it eliminates the startup phase in which agent travel to their equally spaced positions along the cycle}).
\onlyInShortVersion{
Experimental results appear in Figure~\ref{fig:HighDensityTall}-\ref{fig:LowDensityShort}. Refer to \cite{arxive_paper} for additional results. We discuss our results in the next section. 
}
\onlyInLongVersion{
Experimental results appear in Figures~\ref{fig:HighDensityTall}-\ref{fig:MixedDensityMixed}. We discuss our results in the next section. 
}

\newcommand{\coveragePlotTitle}{\text{\small Coverage}}
\newcommand{\numVisitsPlotTitle}{\text{\small Number of Visits}}
\newcommand{\consecutiveVisitsPlotTitle}{\text{\small Ave. Duration Between Visits}}
\newcommand{\timeSpentVisitsPlotTitle}{\text{\small Ave. Time Near Each Point}}

\newcommand{\numAgentsLabel}{\text{\small Number of Agents}}
\newcommand{\timeStepLabel}{\text{\small Time Step}}

\newcommand{\coverageLabel}{\rotatebox{90}{\text{\small Area Covered (\%)}}}
\newcommand{\numVisitsLabel}{\rotatebox{90}{\text{\small Number of visits}}}
\newcommand{\successiveVisitsLabel}{\rotatebox{90}{\text{\small Duration (s)}}}
\newcommand{\timeSpentLabel}{\rotatebox{90}{\text{\small Cumulative Time (s)}}}

\newcommand{\leftFigureMetaTitle}{{\small At 15000 Time Steps}}
\newcommand{\rightFigureMetaTitle}{{\small For Teams of 10 Agents}}

  \newcommand{\widthA}{4.55cm}
  \newcommand{\widthAinside}{4.25cm}
  \newcommand{\heightAinside}{3.2cm}
  \newcommand{\heightSpacerA}{.2cm}

  \begin{figure*}[t!]
  
      \centering
        { High Density Tall Buildings }
  
    \vspace{-.1cm}
  
    \begin{minipage}[t]{0.49\textwidth}   
    
    \centering \leftFigureMetaTitle
    
    \begin{minipage}[t]{1.1\textwidth}
    \begin{minipage}{\widthA}
      \begin{xy}
        \xyimport(100,100){\includegraphics[width=\widthAinside, height=\heightAinside]{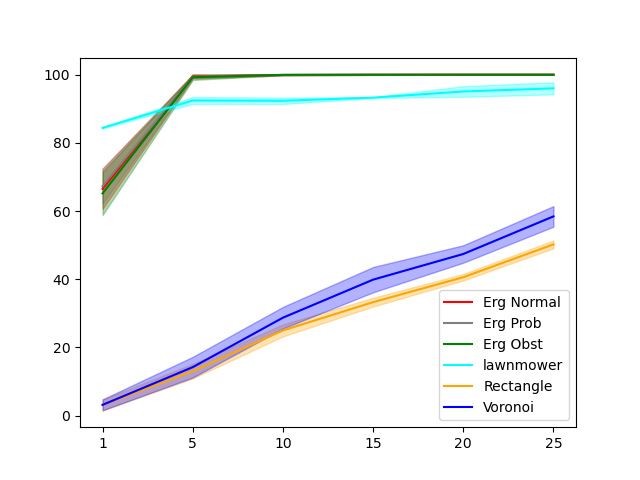}}
               ,(50,95)*{\coveragePlotTitle}
               ,(50,00)*{\numAgentsLabel} 
               ,(4,50)*{\coverageLabel}
      \end{xy}
    \end{minipage}
    \begin{minipage}{\widthA}
      \begin{xy}
        \xyimport(100,100){\includegraphics[width=\widthAinside, height=\heightAinside]{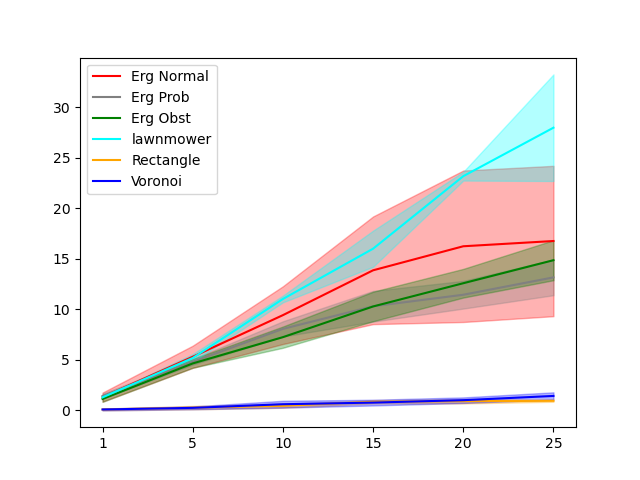}}
               ,(50,95)*{\numVisitsPlotTitle}
               ,(50,00)*{\numAgentsLabel} 
               ,(4,50)*{\numVisitsLabel}
      \end{xy}
    \end{minipage}
    
    \vspace{\heightSpacerA}
    
    \begin{minipage}{\widthA}
      \begin{xy}
        \xyimport(100,100){\includegraphics[width=\widthAinside, height=\heightAinside,  ]{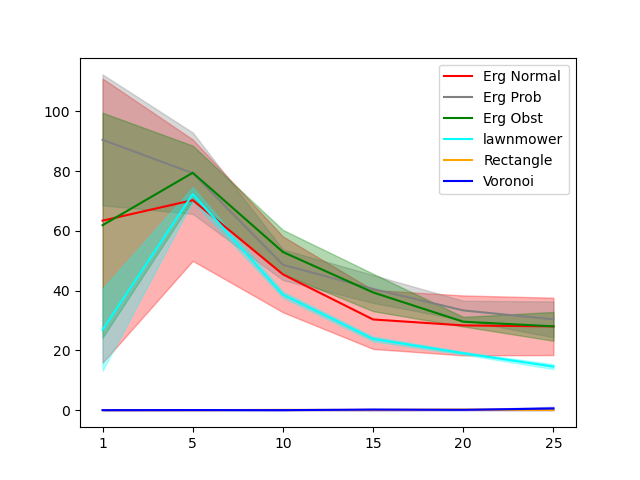}}
               ,(50,95)*{\consecutiveVisitsPlotTitle}
               ,(50,00)*{\numAgentsLabel} 
               ,(4,50)*{\successiveVisitsLabel}
      \end{xy}
    \end{minipage}
    \begin{minipage}{\widthA}
      \begin{xy}
        \xyimport(100,100){\includegraphics[width=\widthAinside, height=\heightAinside,  ]{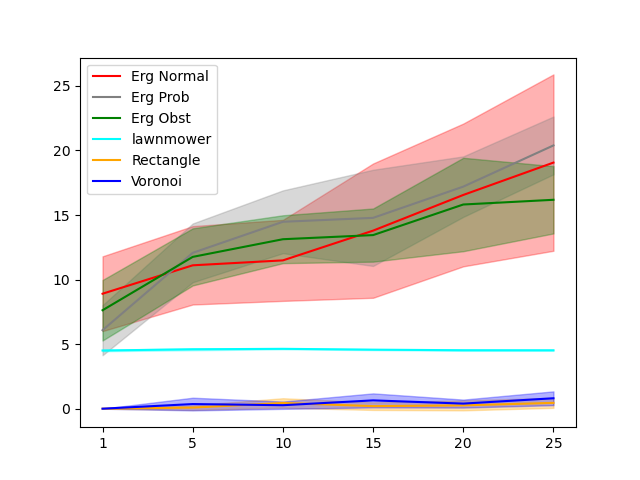}}
               ,(50,95)*{\timeSpentVisitsPlotTitle}
               ,(50,00)*{\numAgentsLabel} 
               ,(4,50)*{\timeSpentLabel}
      \end{xy}
    \end{minipage}
    \end{minipage}
    \end{minipage}
  \begin{minipage}[t]{0.49\textwidth}   
  
      \centering \rightFigureMetaTitle
  
    \begin{minipage}[t]{1.1\textwidth}
    \begin{minipage}{\widthA}
      \begin{xy}
        \xyimport(100,100){\includegraphics[width=\widthAinside, height=\heightAinside,  ]{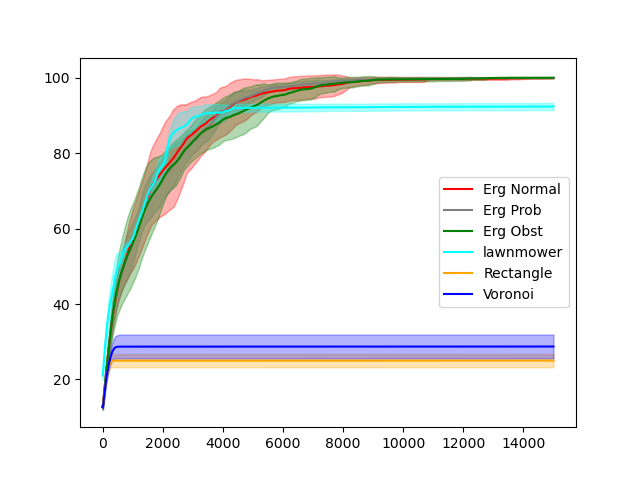}}
               ,(50,95)*{\coveragePlotTitle}
               ,(50,00)*{\timeStepLabel} 
               ,(4,50)*{\coverageLabel}
      \end{xy}
    \end{minipage}
    \begin{minipage}{\widthA}
      \begin{xy}
        \xyimport(100,100){\includegraphics[width=\widthAinside, height=\heightAinside,  ]{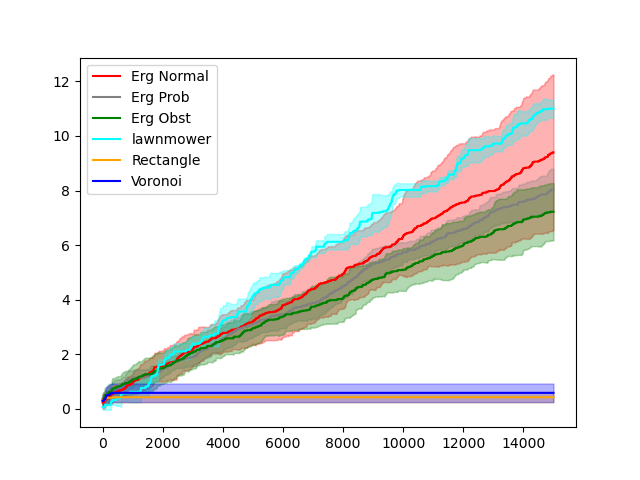}}
               ,(50,95)*{\numVisitsPlotTitle}
               ,(50,00)*{\timeStepLabel} 
               ,(4,50)*{\numVisitsLabel}
      \end{xy}
    \end{minipage}
    
    \begin{minipage}{\widthA}
      \begin{xy}
        \xyimport(100,100){\includegraphics[width=\widthAinside, height=\heightAinside,  ]{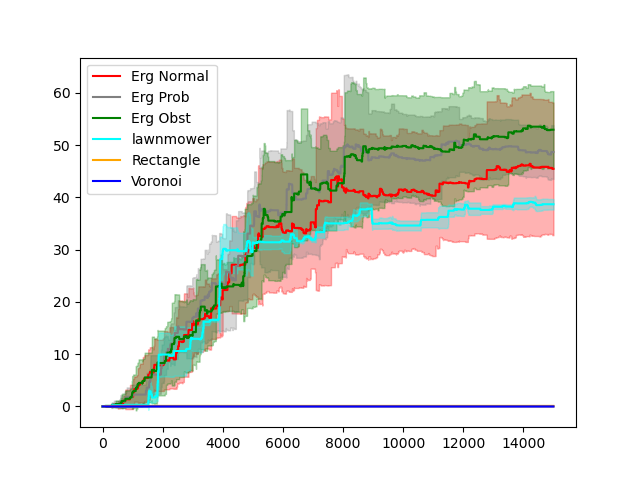}}
               ,(50,95)*{\consecutiveVisitsPlotTitle}
               ,(50,00)*{\timeStepLabel} 
               ,(4,50)*{\successiveVisitsLabel}
      \end{xy}
    \end{minipage}
    \begin{minipage}{\widthA}
      \begin{xy}
        \xyimport(100,100){\includegraphics[width=\widthAinside, height=\heightAinside,  ]{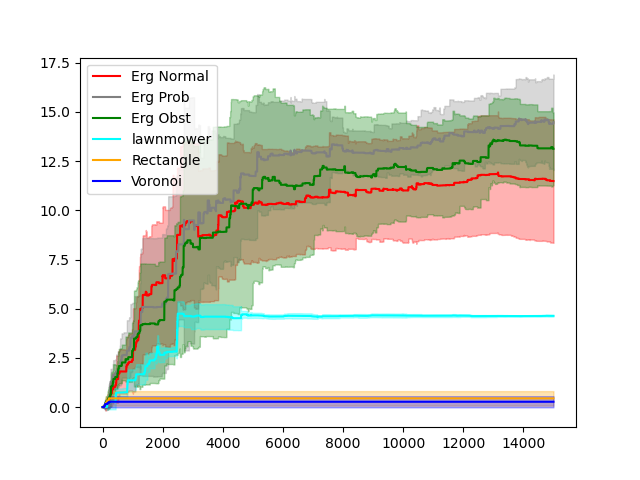}}
               ,(50,95)*{\timeSpentVisitsPlotTitle}
               ,(50,00)*{\timeStepLabel} 
               ,(4,50)*{\timeSpentLabel}
      \end{xy}
    \end{minipage}
    \end{minipage}
    \end{minipage}
 
    

\vspace{.3cm}

  
        \centering
        { Low Density Tall Buildings} 
  
      \vspace{-.1cm}
  
  \begin{minipage}[t]{0.49\textwidth}   
  
      \centering \leftFigureMetaTitle
  
    \begin{minipage}[t]{1.1\textwidth}
    \begin{minipage}{\widthA}
      \begin{xy}
        \xyimport(100,100){\includegraphics[width=\widthAinside, height=\heightAinside,  ]{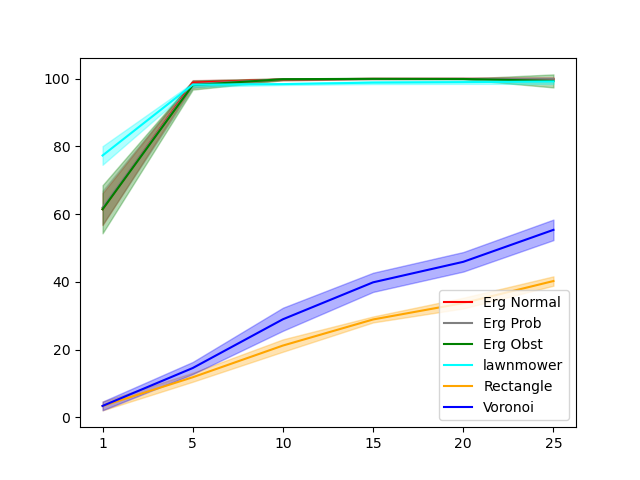}}
               ,(50,95)*{\coveragePlotTitle}
               ,(50,00)*{\numAgentsLabel} 
               ,(4,50)*{\coverageLabel}
      \end{xy}
    \end{minipage}
    \begin{minipage}{\widthA}
      \begin{xy}
        \xyimport(100,100){\includegraphics[width=\widthAinside, height=\heightAinside,  ]{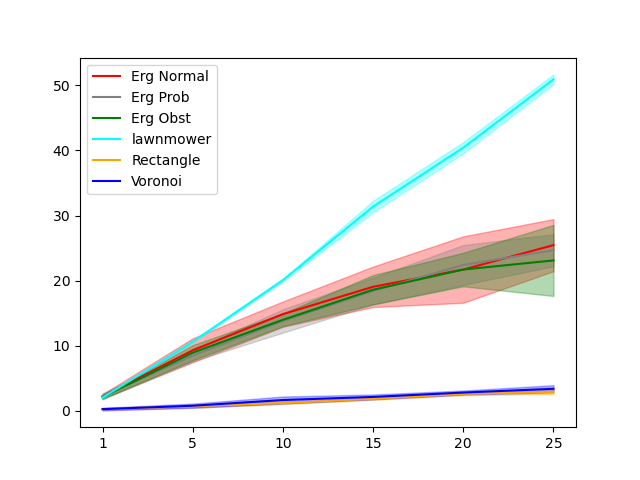}}
               ,(50,95)*{\numVisitsPlotTitle}
               ,(50,00)*{\numAgentsLabel} 
               ,(4,50)*{\numVisitsLabel}
      \end{xy}
    \end{minipage}
    
    \begin{minipage}{\widthA}
      \begin{xy}
        \xyimport(100,100){\includegraphics[width=\widthAinside, height=\heightAinside,  ]{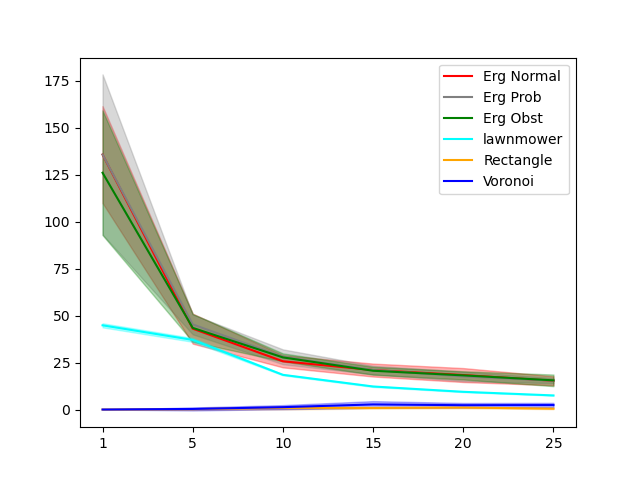}}
               ,(50,95)*{\consecutiveVisitsPlotTitle}
               ,(50,00)*{\numAgentsLabel} 
               ,(4,50)*{\successiveVisitsLabel}
      \end{xy}
    \end{minipage}
    \begin{minipage}{\widthA}
      \begin{xy}
        \xyimport(100,100){\includegraphics[width=\widthAinside, height=\heightAinside,  ]{highdense_m4_graph_1.png}}
               ,(50,95)*{\timeSpentVisitsPlotTitle}
               ,(50,00)*{\numAgentsLabel} 
               ,(4,50)*{\timeSpentLabel}
      \end{xy}
    \end{minipage}
    \end{minipage}
    \end{minipage}
    \begin{minipage}[t]{0.49\textwidth}   
    
        \centering \rightFigureMetaTitle
    
    \begin{minipage}[t]{1.1\textwidth}
    \begin{minipage}{\widthA}
      \begin{xy}
        \xyimport(100,100){\includegraphics[width=\widthAinside, height=\heightAinside,  ]{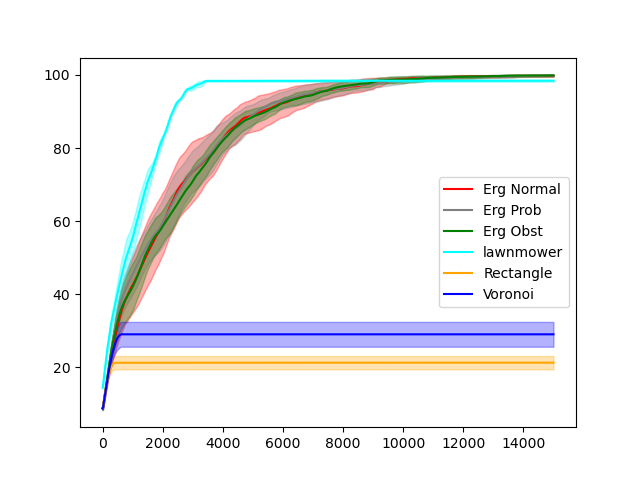}}
               ,(50,95)*{\coveragePlotTitle}
               ,(50,00)*{\timeStepLabel} 
               ,(4,50)*{\coverageLabel}
      \end{xy}
    \end{minipage}
    \begin{minipage}{\widthA}
      \begin{xy}
        \xyimport(100,100){\includegraphics[width=\widthAinside, height=\heightAinside,  ]{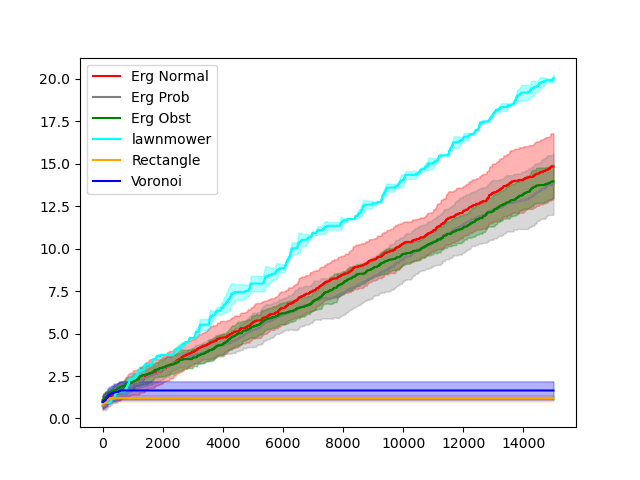}}
               ,(50,95)*{\numVisitsPlotTitle}
               ,(50,00)*{\timeStepLabel} 
               ,(4,50)*{\numVisitsLabel}
      \end{xy}
    \end{minipage}
    
    \begin{minipage}{\widthA}
      \begin{xy}
        \xyimport(100,100){\includegraphics[width=\widthAinside, height=\heightAinside,  ]{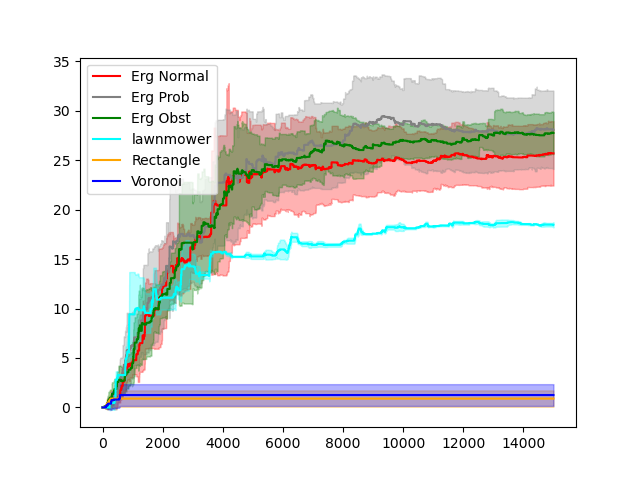}}
               ,(50,95)*{\consecutiveVisitsPlotTitle}
               ,(50,00)*{\timeStepLabel} 
               ,(4,50)*{\successiveVisitsLabel}
      \end{xy}
    \end{minipage}
    \begin{minipage}{\widthA}
      \begin{xy}
        \xyimport(100,100){\includegraphics[width=\widthAinside, height=\heightAinside,  ]{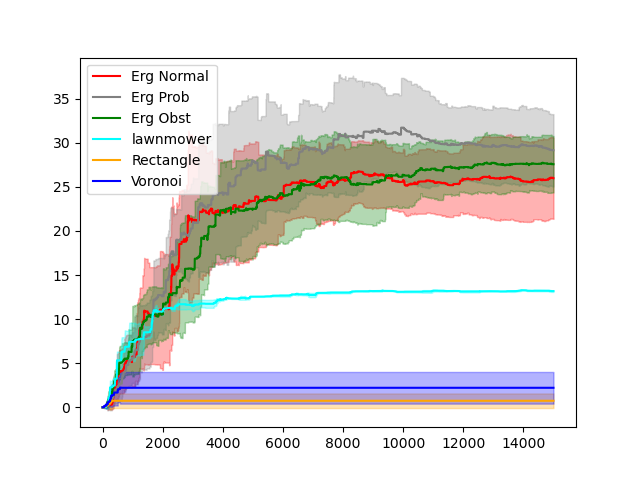}}
               ,(50,95)*{\timeSpentVisitsPlotTitle}
               ,(50,00)*{\timeStepLabel} 
               ,(4,50)*{\timeSpentLabel}
      \end{xy}
    \end{minipage}
    \end{minipage}
    \end{minipage}
    
    
    \caption{Performance of different size teams after 15000 time steps (left) and performance for 10 agents over time (right) in environments with tall buildings and high building density (top 4 rows) or low building density (bottom 4 rows) density.}\label{fig:HighDensityTall}

  \end{figure*}

  \begin{figure*}[t!]
  
          \centering
        { High Density Short Buildings}
  
  \vspace{-.1cm}
  
    \begin{minipage}[t]{0.49\textwidth}   
    
        \centering \leftFigureMetaTitle
    
    \begin{minipage}[t]{1.1\textwidth}
    \begin{minipage}{\widthA}
      \begin{xy}
        \xyimport(100,100){\includegraphics[width=\widthAinside, height=\heightAinside,  ]{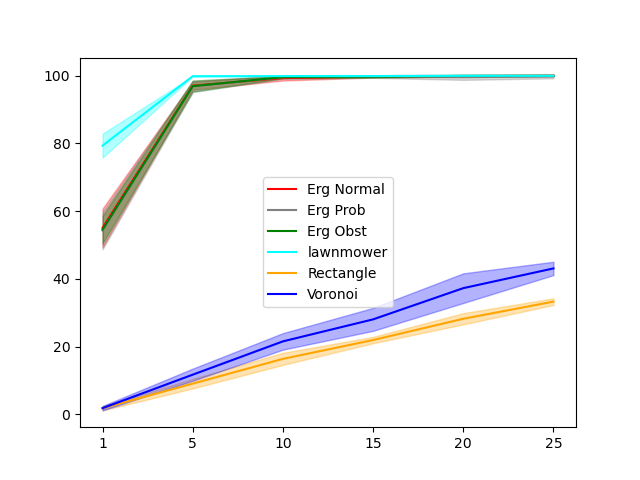}}
               ,(50,95)*{\coveragePlotTitle}
               ,(50,00)*{\numAgentsLabel} 
               ,(4,50)*{\coverageLabel}
      \end{xy}
    \end{minipage}
    \begin{minipage}{\widthA}
      \begin{xy}
        \xyimport(100,100){\includegraphics[width=\widthAinside, height=\heightAinside,  ]{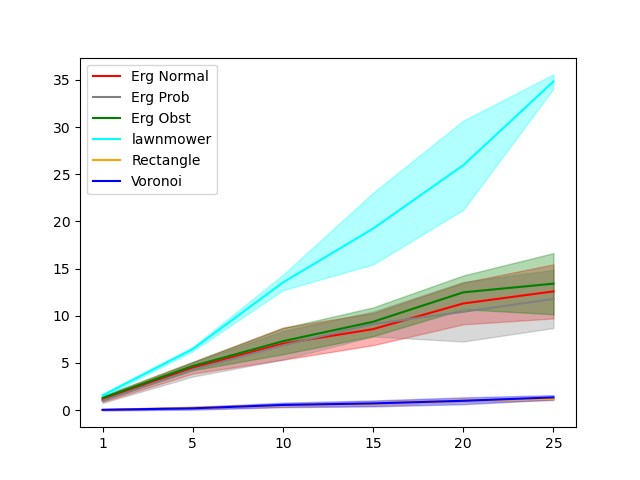}}
               ,(50,95)*{\numVisitsPlotTitle}
               ,(50,00)*{\numAgentsLabel} 
               ,(4,50)*{\numVisitsLabel}
      \end{xy}
    \end{minipage}
    
    \begin{minipage}{\widthA}
      \begin{xy}
        \xyimport(100,100){\includegraphics[width=\widthAinside, height=\heightAinside,  ]{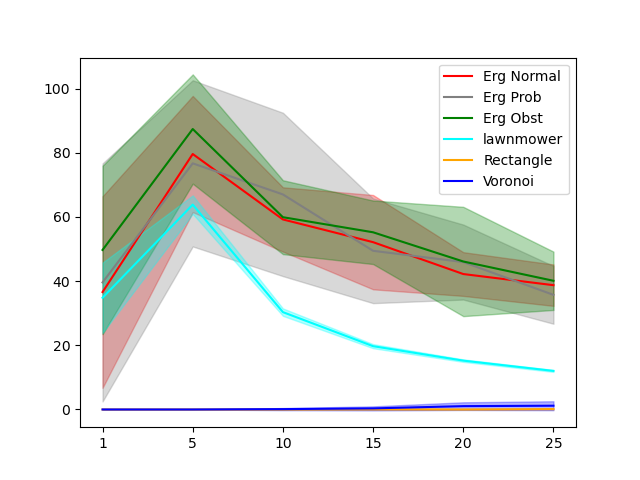}}
               ,(50,95)*{\consecutiveVisitsPlotTitle}
               ,(50,00)*{\numAgentsLabel} 
               ,(4,50)*{\successiveVisitsLabel}
      \end{xy}
    \end{minipage}
    \begin{minipage}{\widthA}
      \begin{xy}
        \xyimport(100,100){\includegraphics[width=\widthAinside, height=\heightAinside,  ]{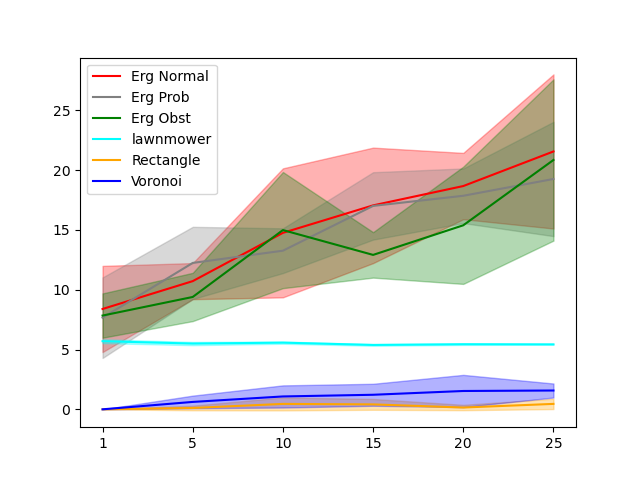}}
               ,(50,95)*{\timeSpentVisitsPlotTitle}
               ,(50,00)*{\numAgentsLabel} 
               ,(4,50)*{\timeSpentLabel}
      \end{xy}
    \end{minipage}
    \end{minipage}
    \end{minipage}
    \begin{minipage}[t]{0.49\textwidth}   
    
        \centering \rightFigureMetaTitle
    
    \begin{minipage}[t]{1.1\textwidth}
    \begin{minipage}{\widthA}
      \begin{xy}
        \xyimport(100,100){\includegraphics[width=\widthAinside, height=\heightAinside,  ]{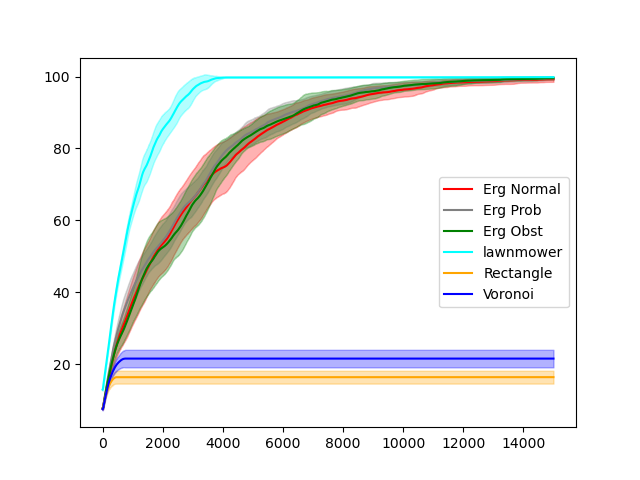}}
               ,(50,95)*{\coveragePlotTitle}
               ,(50,00)*{\timeStepLabel} 
               ,(4,50)*{\coverageLabel}
      \end{xy}
    \end{minipage}
    \begin{minipage}{\widthA}
      \begin{xy}
        \xyimport(100,100){\includegraphics[width=\widthAinside, height=\heightAinside,  ]{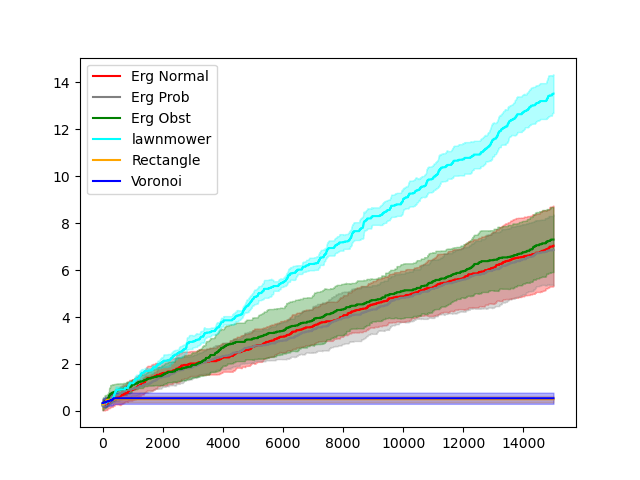}}
               ,(50,95)*{\numVisitsPlotTitle}
               ,(50,00)*{\timeStepLabel} 
               ,(4,50)*{\numVisitsLabel}
      \end{xy}
    \end{minipage}
    
    \begin{minipage}{\widthA}
      \begin{xy}
        \xyimport(100,100){\includegraphics[width=\widthAinside, height=\heightAinside,  ]{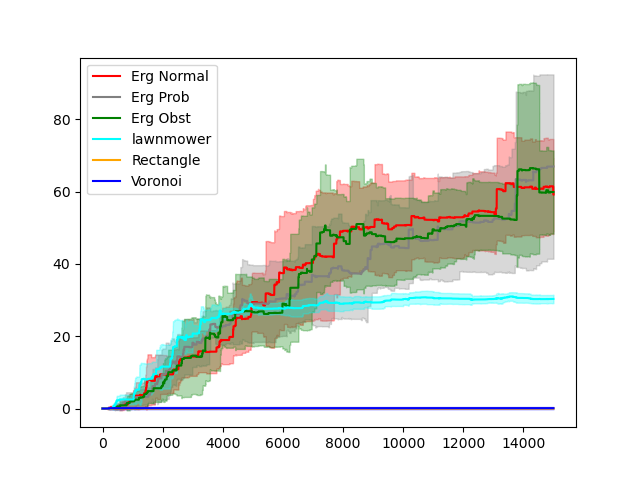}}
               ,(50,95)*{\consecutiveVisitsPlotTitle}
               ,(50,00)*{\timeStepLabel} 
               ,(4,50)*{\successiveVisitsLabel}
      \end{xy}
    \end{minipage}
    \begin{minipage}{\widthA}
      \begin{xy}
        \xyimport(100,100){\includegraphics[width=\widthAinside, height=\heightAinside,  ]{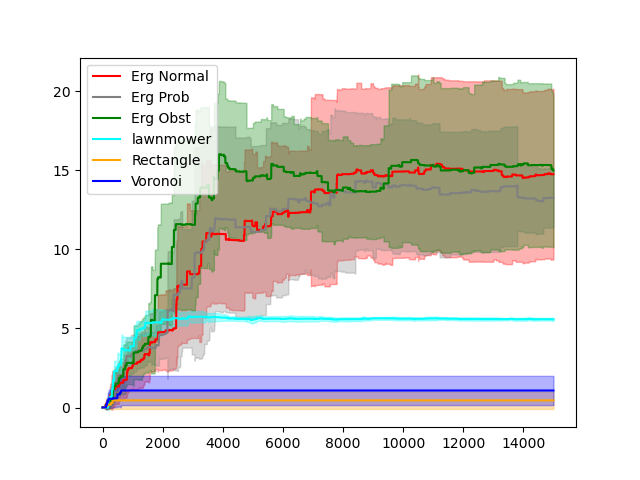}}
               ,(50,95)*{\timeSpentVisitsPlotTitle}
               ,(50,00)*{\timeStepLabel} 
               ,(4,50)*{\timeSpentLabel}
      \end{xy}
    \end{minipage}
    \end{minipage}
    \end{minipage}
    
    \vspace{.3cm}
        
 
  
            \centering
        {Low Density Short Buildings}
  
  \vspace{-.1cm}
  
    \begin{minipage}[t]{0.49\textwidth}   

        \centering \leftFigureMetaTitle
    
    \begin{minipage}[t]{1.1\textwidth}
    \begin{minipage}{\widthA}
      \begin{xy}
        \xyimport(100,100){\includegraphics[width=\widthAinside, height=\heightAinside,  ]{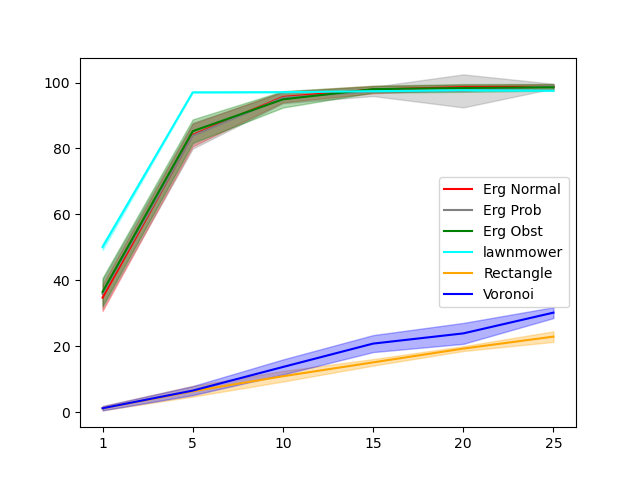}}
               ,(50,95)*{\coveragePlotTitle}
               ,(50,00)*{\numAgentsLabel} 
               ,(4,50)*{\coverageLabel}
      \end{xy}
    \end{minipage}
    \begin{minipage}{\widthA}
      \begin{xy}
        \xyimport(100,100){\includegraphics[width=\widthAinside, height=\heightAinside,  ]{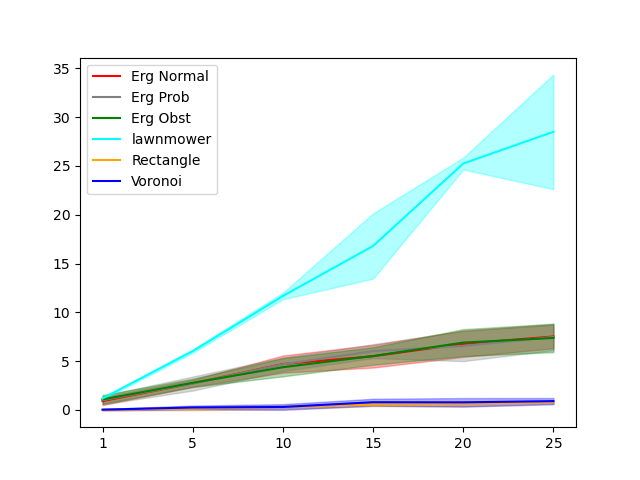}}
               ,(50,95)*{\numVisitsPlotTitle}
               ,(50,00)*{\numAgentsLabel} 
               ,(4,50)*{\numVisitsLabel}
      \end{xy}
    \end{minipage}
    
    \begin{minipage}{\widthA}
      \begin{xy}
        \xyimport(100,100){\includegraphics[width=\widthAinside, height=\heightAinside,  ]{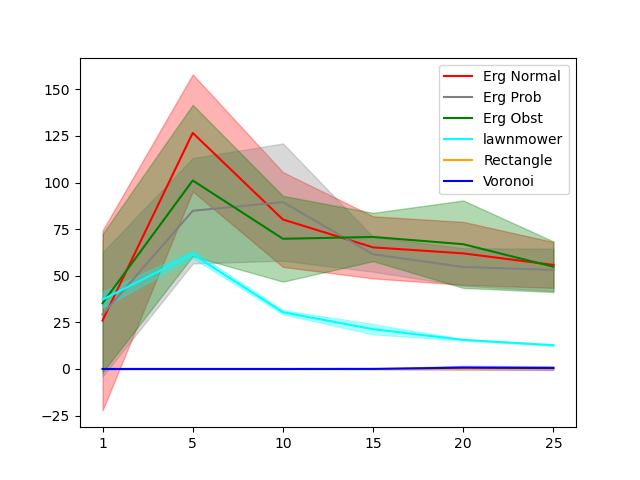}}
               ,(50,95)*{\consecutiveVisitsPlotTitle}
               ,(50,00)*{\numAgentsLabel} 
               ,(4,50)*{\successiveVisitsLabel}
      \end{xy}
    \end{minipage}
    \begin{minipage}{\widthA}
      \begin{xy}
        \xyimport(100,100){\includegraphics[width=\widthAinside, height=\heightAinside,  ]{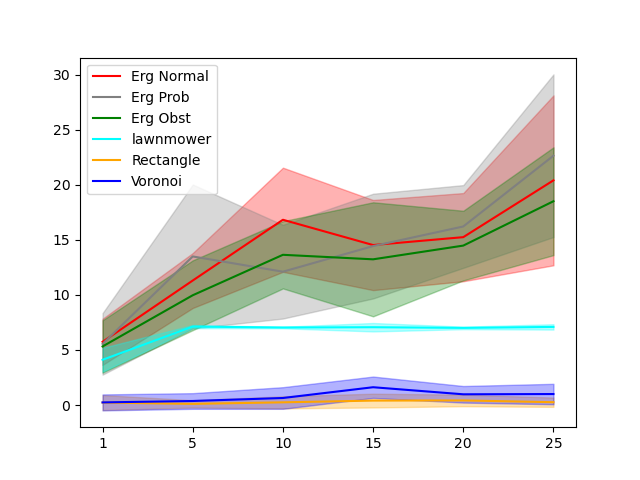}}
               ,(50,95)*{\timeSpentVisitsPlotTitle}
               ,(50,00)*{\numAgentsLabel} 
               ,(4,50)*{\timeSpentLabel}
      \end{xy}
    \end{minipage}
    \end{minipage}
    \end{minipage}
  \begin{minipage}[t]{0.49\textwidth}   
  
      \centering \rightFigureMetaTitle
  
    \begin{minipage}[t]{1.1\textwidth}
    \begin{minipage}{\widthA}
      \begin{xy}
        \xyimport(100,100){\includegraphics[width=\widthAinside, height=\heightAinside,  ]{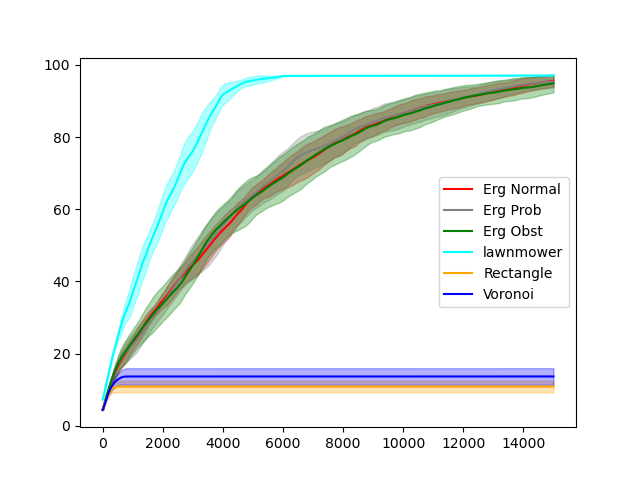}}
               ,(50,95)*{\coveragePlotTitle}
               ,(50,00)*{\timeStepLabel} 
               ,(4,50)*{\coverageLabel}
      \end{xy}
    \end{minipage}
    \begin{minipage}{\widthA}
      \begin{xy}
        \xyimport(100,100){\includegraphics[width=\widthAinside, height=\heightAinside,  ]{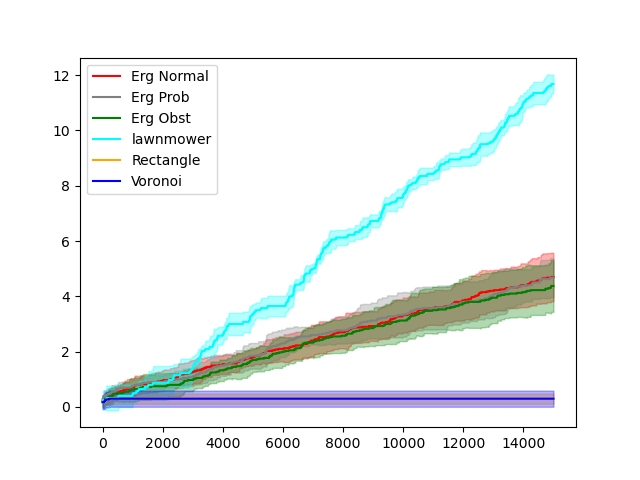}}
               ,(50,95)*{\numVisitsPlotTitle}
               ,(50,00)*{\timeStepLabel} 
               ,(4,50)*{\numVisitsLabel}
      \end{xy}
    \end{minipage}
    
    \begin{minipage}{\widthA}
      \begin{xy}
        \xyimport(100,100){\includegraphics[width=\widthAinside, height=\heightAinside,  ]{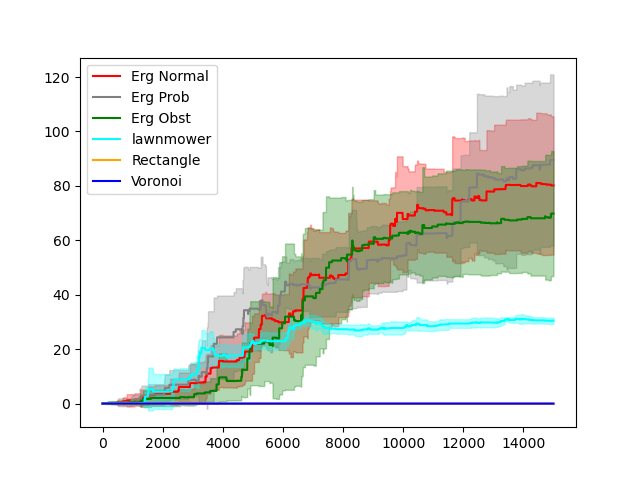}}
               ,(50,95)*{\consecutiveVisitsPlotTitle}
               ,(50,00)*{\timeStepLabel} 
               ,(4,50)*{\successiveVisitsLabel}
      \end{xy}
    \end{minipage}
    \begin{minipage}{\widthA}
      \begin{xy}
        \xyimport(100,100){\includegraphics[width=\widthAinside, height=\heightAinside,  ]{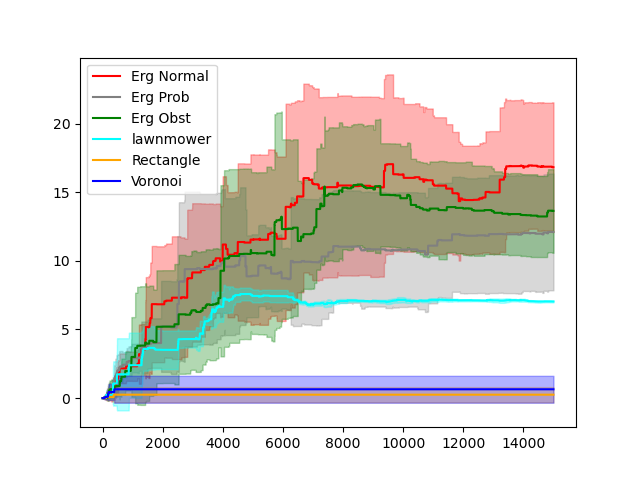}}
               ,(50,95)*{\timeSpentVisitsPlotTitle}
               ,(50,00)*{\timeStepLabel} 
               ,(4,50)*{\timeSpentLabel}
      \end{xy}
    \end{minipage}
    \end{minipage}
    \end{minipage}
    
  
      \caption{Performance of different size teams after 15000 time steps (left) and performance for 10 agents over time (right) in environments with short buildings and high building density (top 4 rows) or low building density (bottom 4 rows) density.}\label{fig:LowDensityShort}

  \end{figure*}

\onlyInLongVersion{

  \begin{figure*}[t!]
            \centering
        {\large Mixed Density Mixed Height Buildings}
  
    \begin{minipage}[t]{0.49\textwidth}   
    
    \centering \leftFigureMetaTitle
    
    \begin{minipage}[t]{1.1\textwidth}
    \begin{minipage}{\widthA}
      \begin{xy}
        \xyimport(100,100){\includegraphics[width=\widthAinside, height=\heightAinside,  ]{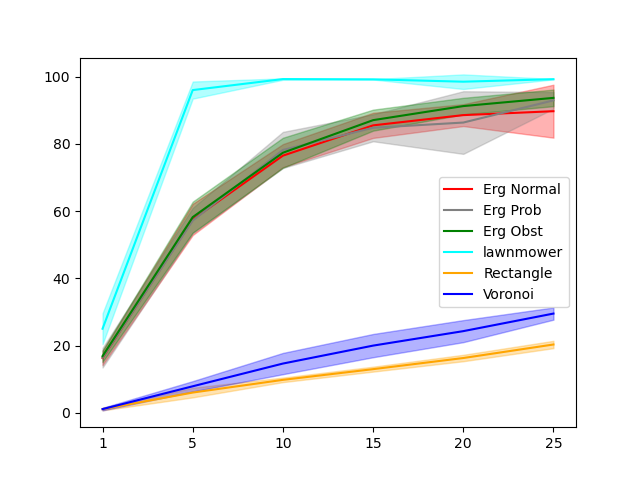}}
               ,(50,95)*{\coveragePlotTitle}
               ,(50,00)*{\numAgentsLabel} 
               ,(4,50)*{\coverageLabel}
      \end{xy}
    \end{minipage}
    \begin{minipage}{\widthA}
      \begin{xy}
        \xyimport(100,100){\includegraphics[width=\widthAinside, height=\heightAinside,  ]{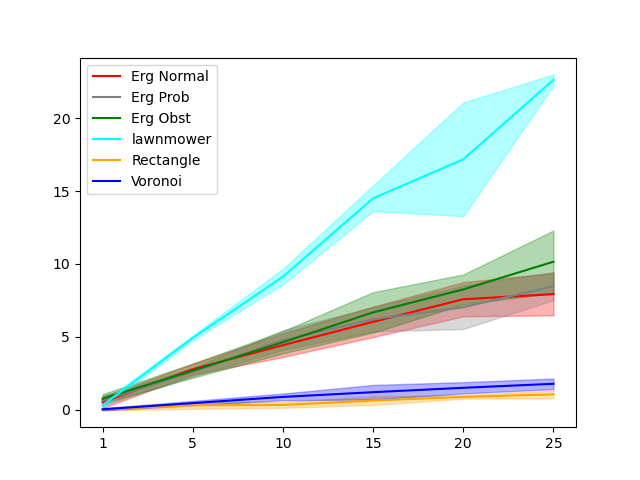}}
               ,(50,95)*{\numVisitsPlotTitle}
               ,(50,00)*{\numAgentsLabel} 
               ,(4,50)*{\numVisitsLabel}
      \end{xy}
    \end{minipage}
    
    \begin{minipage}{\widthA}
      \begin{xy}
        \xyimport(100,100){\includegraphics[width=\widthAinside, height=\heightAinside,  ]{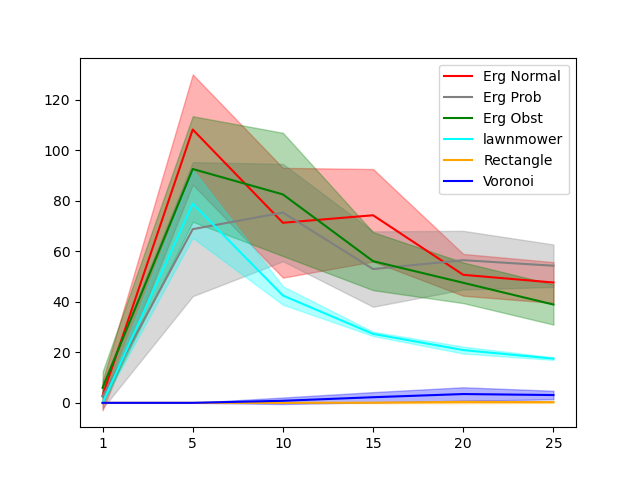}}
               ,(50,95)*{\consecutiveVisitsPlotTitle}
               ,(50,00)*{\numAgentsLabel} 
               ,(4,50)*{\successiveVisitsLabel}
      \end{xy}
    \end{minipage}
    \begin{minipage}{\widthA}
      \begin{xy}
        \xyimport(100,100){\includegraphics[width=\widthAinside, height=\heightAinside,  ]{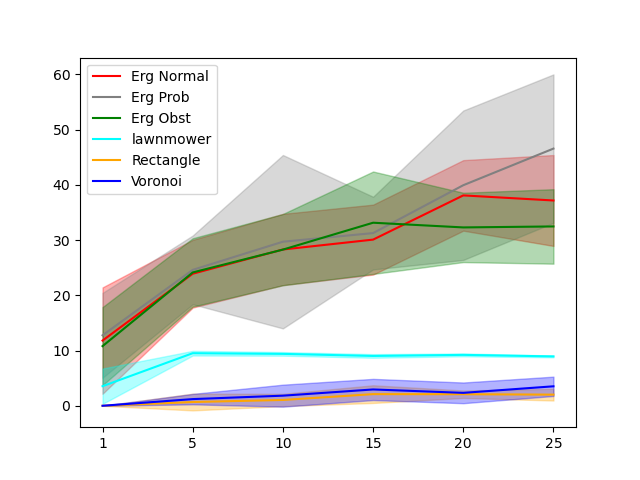}}
               ,(50,95)*{\timeSpentVisitsPlotTitle}
               ,(50,00)*{\numAgentsLabel} 
               ,(4,50)*{\timeSpentLabel}
      \end{xy}
    \end{minipage}
    \end{minipage}
    \end{minipage}
      \begin{minipage}[t]{0.49\textwidth}   
      
          \centering \rightFigureMetaTitle
      
    \begin{minipage}[t]{1.1\textwidth}
    \begin{minipage}{\widthA}
      \begin{xy}
        \xyimport(100,100){\includegraphics[width=\widthAinside, height=\heightAinside,  ]{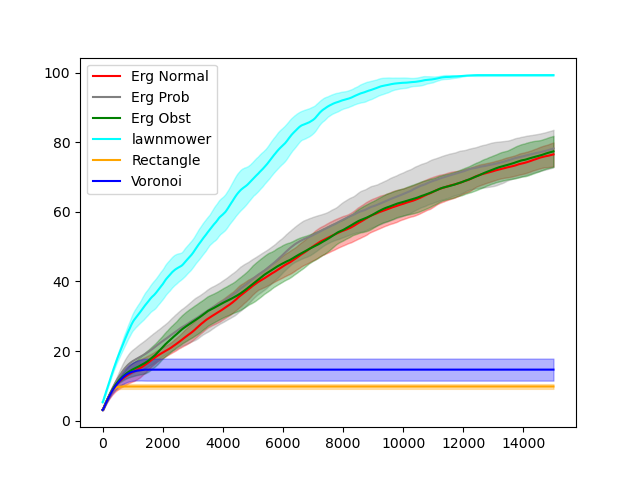}}
               ,(50,95)*{\coveragePlotTitle}
               ,(50,00)*{\timeStepLabel} 
               ,(4,50)*{\coverageLabel}
      \end{xy}
    \end{minipage}
    \begin{minipage}{\widthA}
      \begin{xy}
        \xyimport(100,100){\includegraphics[width=\widthAinside, height=\heightAinside,  ]{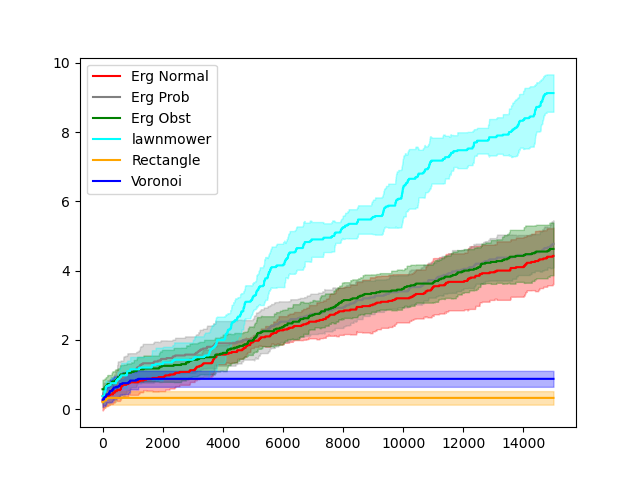}}
               ,(50,95)*{\numVisitsPlotTitle}
               ,(50,00)*{\timeStepLabel} 
               ,(4,50)*{\numVisitsLabel}
      \end{xy}
    \end{minipage}
    
    \begin{minipage}{\widthA}
      \begin{xy}
        \xyimport(100,100){\includegraphics[width=\widthAinside, height=\heightAinside,  ]{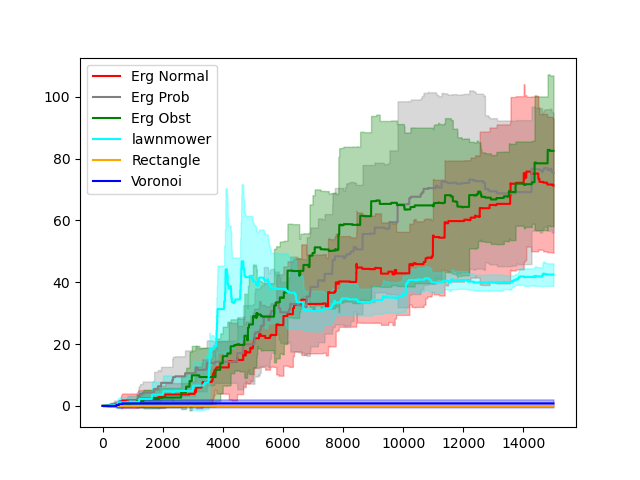}}
               ,(50,95)*{\consecutiveVisitsPlotTitle}
               ,(50,00)*{\timeStepLabel} 
               ,(4,50)*{\successiveVisitsLabel}
      \end{xy}
    \end{minipage}
    \begin{minipage}{\widthA}
      \begin{xy}
        \xyimport(100,100){\includegraphics[width=\widthAinside, height=\heightAinside,  ]{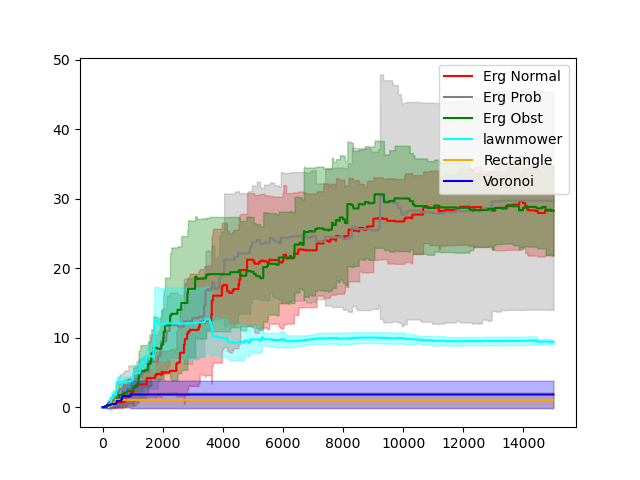}}
               ,(50,95)*{\timeSpentVisitsPlotTitle}
               ,(50,00)*{\timeStepLabel} 
               ,(4,50)*{\timeSpentLabel}
      \end{xy}
    \end{minipage}
    \end{minipage}
    \end{minipage}
    \caption{Performance of different size teams (left) and performance over time (right) in environments with mixed density and mixed buildings.}\label{fig:MixedDensityMixed}
    
  \end{figure*}

















}

\section{Results} \label{sec:results}



In tall environments, Algorithms that consider urban obstacles tend to have better performance than those that do not. This makes sense because more time is required to fly up and over buildings than around them. Biasing movement toward the free space (but still flying over them when necessary) outperforms explicit obstacle avoidance in {\it short} environments. Building height effects become more pronounced as team size increases. 
%
%
Increasing building density tends to amplify the differences between the different methods.





The Ergodic methods revisit a particular point less often, on average, than lawnmower sweep.
\extraDiscussion{In practice, there are adversarial situations in which the benefits of non-deterministic revisit times are preferred, even if this causes an increase in expected revisit time.}
The obstacle avoiding Ergodic has lower revisit times than free space biased ergodic, which has lower revisit times than normal Ergodic.


Our experiments provide insights into the trade-off between algorithms that are designed for static and dynamic cases. The percentage of the environment that is covered by the Voronoi and grid methods appears to have a linear relationship with agent number, and an approximately linear function of time for lawn mower sweep and the Ergodic methods (until a maximum value is reached).

Using the Voronoi and Grid methods, the average time between visits is $0$, i.e., instantaneous---at least, to those points that are ever visited.  This reflects the fact that methods designed for static coverage have agents move to and remain at static locations. The lawn-mower and Ergodic methods have decreasing average revisit times with more agents, as well as increasing mean and standard deviations of revisit times as time increases. The increasing mean revisit time is partially an artifact of the fact that we initialize revisit times to 0 (since re-visit implies at least two visits, and so a point is essentially not tracked, with respect to this metric, until it is visited a second time). The increasing standard deviation appears to asymptote toward a constant value for the lawn mower algorithms, but continues grow over the timescales that we evaluate for the stochastic algorithms. 
Similar results are reflected in the total visit count to each revisited point (lawn mower and Ergodic continue to increase, while Voronoi and grid method taper off).


%
%
%
%

\section{Summary and Conclusion} \label{sec:conclusions}

The main contribution of this work is an empirical evaluation of six multi-agent coverage algorithm in urban environments. Urban environments present interesting challenges for coverage algorithms because buildings and other structures are obstacles to navigation. Such obstacles are especially relevant when the coverage sensor cannot be used above a maximum altitude and that altitude is below the tallest building in the environment.

We evaluate two static and four dynamic coverage algorithms. Three of the dynamic coverage algorithms are relevant to adversarial scenarios and use ergodicity to create random revisit times to each location in the environment. These differ in their approach to obstacle avoidance. Biasing ergodic motion away from obstacles (and then greedily flying over buildings in the rare cases it is necessary), appears to work well in environments with short buildings. On the other hand, it appears better to maneuver around tall buildings. While this is an intuitive result, previous ergodic methods appearing in the literature consider only 2D coverage scenarios. Our paper investigates how such algorithms perform when obstacles are 3D and the coverage region is defined to be the 2D ground plane.




%

Our results also confirm the intuition that, for the dynamic methods, mean revisit times appear to scale linearly with agent number. While for the static methods, coverage percentage appear to scale linearly with agent number.
%
We observe that the grid-based method occasionally outperformed the Voronoi method with respect to the time spent in sampled areas, especially for teams with $\leq15$ agents.

%





\bibliographystyle{IEEEtran}
\bibliography{main}

\end{document}